\documentclass[screen]{jair}

\usepackage{bm}
\usepackage{bbm}
\usepackage{algorithmic}
\usepackage{algorithm}
\usepackage{amsthm}
\usepackage{caption}

\newtheorem{theorem}{Theorem}

\newtheorem{lemma}{Lemma}

\newtheorem{proposition}{Proposition}

\newcommand{\revision}[1]{\textcolor{black}{#1}} 

\setcopyright{cc}
\copyrightyear{2025}
\acmYear{2025}
\acmDOI{10.1613/jair.1.17526}

\JAIRAE{Scott Sanner}
\JAIRTrack{}
\acmVolume{83}
\acmArticle{4}
\acmMonth{6}
\acmYear{2025}
\begin{document}

%\title{JAIR Example Template}
\title[DSAC: Distributional Soft Actor-Critic]{DSAC: Distributional Soft Actor-Critic for Risk-Sensitive Reinforcement Learning}

%% used to denote shared contribution to the research and/or corresponding author.

\author{Xiaoteng Ma}
\authornote{Equal contribution.}
\orcid{0000-0002-7250-6268}
\email{pony.xtma@gmail.com}
% \orcid{https://orcid.org/0000-0002-2037-3694} % This works too if you want to see the full URL
\affiliation{%
	\institution{Department of Automation, Tsinghua University}
	\city{Beijing}
	\country{China}
}

\author{Junyao Chen}
\authornotemark[1]
\orcid{0009-0002-5849-2020}
\email{junyao.chen@columbia.edu}
\affiliation{%
	\institution{School of Engineering and Applied Science, Columbia University}
	\city{New York}
	\state{NY}
	\country{United States}}

\author{Li Xia}
\authornote{Corresponding author.}
\orcid{0000-0001-9141-2569}
\email{xiali5@sysu.edu.cn}
\affiliation{%
	\institution{School of Business, Sun Yat-sen University}
	\city{Guangzhou}
	\state{Guangdong}
	\country{China}
}

\author{Jun Yang}
\orcid{0000-0002-9386-5825}
\email{yangjun603@tsinghua.edu.cn}
\affiliation{%
	\institution{Department of Automation, Tsinghua University}
	\city{Beijing}
	\country{China}
}

\author{Qianchuan Zhao}
\orcid{0000-0002-7952-5621}
\email{zhaoqc@tsinghua.edu.cn}
\affiliation{%
	\institution{Department of Automation, Tsinghua University}
	\city{Beijing}
	\country{China}
}

\author{Zhengyuan Zhou}
\orcid{0000-0002-0005-9411}
\email{zzhou@stern.nyu.edu}
\affiliation{%
	\institution{Stern School of Business, New York University}
	\city{New York}
	\state{NY}
	\country{United States}
}

\renewcommand{\shortauthors}{Ma, Chen, et al.}

\begin{abstract}
	We present Distributional Soft Actor-Critic (DSAC), a distributional reinforcement learning (RL) algorithm that combines the strengths of distributional information of accumulated rewards and entropy-driven exploration from Soft Actor-Critic (SAC) algorithm. DSAC models the randomness in both action and rewards, surpassing baseline performances on various continuous control tasks. Unlike standard approaches that solely maximize expected rewards, we propose a unified framework for risk-sensitive learning, one that optimizes the risk-related objective while balancing entropy to encourage exploration. Extensive experiments demonstrate DSAC's effectiveness in enhancing agent performances for both risk-neutral and risk-sensitive control tasks.
\end{abstract}

%\begin{abstract}
%      A clear and well-documented \LaTeX\ document is presented as an
%  article formatted for publication by ACM in a conference proceedings
%  or journal publication. Based on the ``acmart'' document class, this
%  article presents and explains many of the common variations, as well
%  as many of the formatting elements an author may use in the
%  preparation of the documentation of their work.
%\end{abstract}

%% JAIR Note: 
%% Do not include ACM CCS Concepts or Keywords

%% To be updated by authors.
\received{1 November 2022}
\received[revised]{3 November 2024}
\received[accepted]{28 April 2025}

\maketitle

\section{Introduction}
\label{sec:intro}
In the past few years, model-free deep reinforcement learning~\cite{sutton2018reinforcement} has been a powerful and applicable paradigm ~\cite{mnih2013playing,silver2016mastering,gu2017deep}. Such well-founded optimism stems from the substantial performance gains that deep neural networks can potentially unlock in reinforcement learning — whether by parameterizing value functions or policies — by leveraging their powerful representational capacity. However, simply throwing in ``deep function approximator'' is far from sufficient. Building effective RL algorithms often requires well-integrating accurate and efficient function approximation.

One important such aspect is exploration: since current actions influence the future state and hence the future rewards of the underlying Markov Decision Process \cite{Puterman1994Markov,bertsekas1995dynamic}, effective exploration is a key aspect of RL algorithms. In model-free RL literature, randomness in action space is widely employed as a primitive to balance exploration and exploitation. On-policy algorithms, such as A3C~\cite{a3c}, TRPO~\cite{trpo} and PPO~\cite{ppo}, use stochastic action space and optimize the parameters by policy gradient. However, on-policy algorithms are not data-efficient as they require new experience for policy evaluation. On the other hand, off-policy methods can reuse past experience and hence improve data efficiency. However, most off-policy methods, such as DQN~\cite{dqn}, DDPG~\cite{ddpg} and TD3~\cite{td3}, inherit the simple $\epsilon$-greedy strategy from Q-learning by adding a small noise to a deterministic policy for exploration, where the scale of action perturbation is hard to choose in practice. To improve robustness of off-policy methods, SVG~\cite{svg} introduces re-parameterized stochastic policy, while its objective is still standard maximum expected discounted returns.

The \emph{maximum entropy} (MaxEnt) RL~\cite{nachum2017bridging,sac} has shown promise for encouraging the randomness (and hence diversity) of actions by formally formulating the entropy of a policy into the objective. This maximum entropy approach is based on theoretical principles and has been applied to inverse reinforcement learning~\cite{meirl,zhou2018infinite} and optimal control~\cite{todorov2008general,rawlik2013stochastic}.
\citet{sac} propose soft Actor-Critic (SAC), a continuous action space algorithm that has achieved superior performance in many continuous action control tasks. SAC computes an optimal policy by minimizing the KL-divergence between the action distribution and the exponential form of the soft action-value function. 
Many works have been done for understanding the effectiveness of the maximum entropy objective~\cite{pgm-rl,schulman2017equivalence,ahmed2019understanding,cen2020fast}, which reveal that objective with entropy regularization enjoys the smoother optimization landscape and the faster convergence rate, and builds connections between RL with probabilistic graphical models and convex optimization.

Concurrently, \textit{distributional RL}~\cite{sobel1982variance,morimura2010nonparametric,c51} considers the whole distribution of value functions, rather than just the expectation, to make more informed decisions that lead to superior rewards. Incidentally, experiments show that similar encouraging mechanisms also exist in human brains~\cite{dabney2020distributional}. 
	To address the challenge of approximating the distribution of value functions, two categories of approaches dominate: learning discrete categorical distributions (CDRL)~\cite{c51,d4pg,rowland2018analysis,qu2019nonlinear,bellemare2019distributional} and learning quantiles of distributions (QDRL)~\cite{qr-dqn,iqn,edrl,zhang2019quota,bodnar2019quantile}. Comparing with CDRL, QDRL learns quantiles of distribution directly without any assumption about the range of return. Moreover, quantiles are naturally related to risk measures. Thus, QDRL shows more potential than CDRL in recent works.

SAC and distributional RL each have limitations: SAC only considers the first moment of values, while distributional RL lacks action diversity for exploration. This leads to the question: \emph{Can we exploit randomness to enhance action diversity and leverage distributional information?} As demonstrated in this paper, the answer is yes. Previous work on integrating of SAC and distributional RL assumes a Gaussian distribution on the return distribution $Z$ for its parametrization~\cite{duan2021distributional}, which may not capture skewed or heavy-tailed returns. Our method uses quantile regression, providing a non-parametric model that better captures asymmetries and tail risks, leading to more robust decision-making under uncertainty.

\subsection{Our Contributions}
Our contributions are threefold. First, we present Distributional Soft Actor-Critic (DSAC), that combines MaxEnt RL with distributional RL, by leveraging the distributional information of value functions in SAC and encouraging action randomness through entropy in distributional RL. We define a distributional soft Bellman operator and use quantile regression to estimate the soft discounted returns for continuous action tasks.

Second, we adapt DSAC for risk-sensitive learning, providing a unified framework that handles multiple risk measures, such as variance~\cite{sobel1982variance,xia2016optimization,fu2018risk}, CVaR~\cite{chow2015risk} and CPT~\cite{cpt}. Aforementioned risk measures all involve higher moments of the underlying value distribution while they do not satisfy the Bellman equation. Prior work~\cite{iqn,singh2020improving} has demonstrated the ability to handle multiple risk measures. Inspired by these foundations, our approach supports multiple metrics and allows for optimizing risk while balancing the entropy of policy, ensuring the exploratory capability and robustness of the policy in risky scenarios.

Third, we perform extensive experiments comparing DSAC with existing model-free RL algorithms in continuous state and actions spaces. DSAC achieves superior performance on standard benchmarks (MuJoCo and Box2d in OpenAI Gym), surpassing SAC, TD3, SDPG, and D4PG. We also demonstrated the effectiveness of our risk-sensitive RL framework through simulation experiments.

\section{Background}
We operate within the context of a standard MDP~\cite{Puterman1994Markov}, characterized by the tuple $\langle \mathcal{S}, \mathcal{A}, R, P, \gamma \rangle$. Here, $\mathcal{S}$ and $\mathcal{A}$ represent the continuous state and action spaces, respectively. The transition probability density from one state to another, given an action, is denoted by $P: \mathcal{S} \times \mathcal{S} \times \mathcal{A} \to [0, \infty)$ and is considered unknown. The reward function is represented by $R: \mathcal{S} \times \mathcal{A} \to \mathbb{R}$, and $\gamma \in (0,1)$ is the discount factor. A policy $\pi:\mathcal{S}\to \mathscr{P} (\mathcal{A})$ is a mapping from each state to a probability distribution over actions in $\mathcal{A}$. The set of all policies is denoted by $\Pi$.
		
	The goal of standard RL is to maximize the expected sum of discounted rewards, given by:
	\begin{equation}
		\label{equ:rl-obj}
		\mathcal{J}(\pi) =  \mathbb{E}_\pi\left[\sum_{t=0}^\infty \gamma^t R\left(s_{t}, a_t\right)\right],
	\end{equation}
	with the initial state $s_0$ distributed according to $d_0(s)$.
		
	For a given policy $\pi \in \Pi$, the \textit{action-value function} $Q^\pi:\mathcal{S} \times \mathcal{A} \to \mathbb{R}$ is defined as:
	\begin{equation*}
		Q^\pi(s, a) := \mathbb{E}_\pi \left[ \sum_{t=0}^\infty \gamma^t R(s_t, a_t) \right],a_t \sim \pi(\cdot \mid s_t), s_{t+1} \sim P(\cdot \mid s_t, a_t), s_0 = s, a_0 = a.
	\end{equation*}
	The \emph{Bellman operator} $\mathcal{T}^\pi$ and the \emph{Bellman optimality operator} $\mathcal{T}^*$ are defined as:
	\begin{equation}
		\begin{split}
			\mathcal{T}^{\pi} Q(s, a) &:= \mathbb{E} \left[R(s, a)\right] + \gamma \mathbb{E}_{P,\pi} [Q\left(s^{\prime}, a^{\prime}\right)], \\
			\mathcal{T}^* Q(s, a) &:= \mathbb{E} \left[R(s, a)\right] + \gamma \max_{a^{\prime}} \mathbb{E}_{P} [Q\left(s^{\prime}, a^{\prime}\right)].
		\end{split}
	\end{equation}
	Applying either operator iteratively from some initial $Q_0$ will converge to its fixed point $Q^\pi$ or $Q^*$ at a geometric rate, as both operators are contractive~\cite{ndp}.

	\subsection{Maximum Entropy Reinforcement Learning}
		
	MaxEnt RL diverges from standard RL by maximizing the sum of rewards while simultaneously maximizing the entropy of the policy. The objective function of MaxEnt RL is given by:
	\begin{equation}
		\label{equ:sac-obj}
		\mathcal{J}(\pi) = \mathbb{E}_\pi \left[ \sum_{t=0}^\infty \gamma^t \left( R\left(s_{t}, a_t\right)+\alpha \mathcal{H}\left(\pi\left(\cdot \mid s_t \right)\right)\right)\right].
	\end{equation}
    where $\alpha$ is a temperature parameter.
	The \textit{soft action-value function} $Q^\pi:\mathcal{S} \times \mathcal{A} \to \mathbb{R}$ for a policy $\pi \in \Pi$ is defined as:
	\begin{equation*}
		\begin{split}
			Q^\pi(s, a) &:= \mathbb{E}_\pi \left[ R(s, a) + \sum_{t=1}^\infty \gamma^t [R(s_t, a_t) - \alpha \log \pi(a_t \mid s_t) ] \right],\\
			&a_t \sim \pi(\cdot \mid s_t), s_{t+1} \sim P(\cdot \mid s_t, a_t), s_0 = s, a_0 = a.    
		\end{split}
	\end{equation*}
	By augmenting the standard Bellman operator $\mathcal{T}^\pi$ with an entropy regularization term, we define the \emph{soft Bellman operator} $\mathcal{T}_{S}^\pi$ and the \emph{soft optimality Bellman operator} $\mathcal{T}_{S}^*$~\cite{nachum2017bridging} as:
	\begin{align}
		\label{equ:soft-op}
		\mathcal{T}_{S}^\pi Q(s, a) & := \mathbb{E} [R]+\gamma \mathbb{E}_{P,\pi}\left[Q(s',a') - \alpha \log \pi (a' \mid s') \right], \notag \\
		\mathcal{T}_{S}^* Q(s, a)   & := \mathbb{E} [R]+\gamma \mathbb{E}_{P}\left[\alpha \log( \|\exp (Q(s',\cdot)/\alpha)\|_1 ) \right].     
	\end{align}
	The soft Bellman operators retain the $\gamma$-contraction property of the original Bellman operators. The key distinction between $\mathcal{T}_{S}^*$ and $\mathcal{T}^*$ is that the unique fixed point of $\mathcal{T}_{S}^*$ corresponds to a unique stochastic policy in the softmax form of $Q^*$: $\pi^*(\cdot \mid s) \propto \exp(Q^*(s, \cdot)/\alpha)$.

	\citet{sac} introduced the Soft Actor-Critic (SAC) algorithm for learning policies in continuous action spaces with a MaxEnt objective function. SAC alternates between soft policy evaluation, implemented by repeatedly applying $\mathcal{T}_{S}^\pi$, and soft policy improvement until convergence is achieved.
		
	The soft policy improvement is realized by minimizing the Kullback-Leibler divergence between the policy distribution and the exponential form of the soft action-value function:
	\begin{equation}
		\label{equ:sac-ori}
		\pi_{\mathrm{new}}=\mathop{\arg\min}_{\pi^{\prime} \in \Pi} \mathrm{D}_{\mathrm{KL}}\left(\pi^{\prime}\left(\cdot \mid s\right) \Vert \frac{ \exp \left(Q^{\pi_{\mathrm{old}}}\left(s, \cdot\right) /\alpha\right)}{\Delta^{\pi_{\mathrm{old}}}\left(s\right)}\right),
	\end{equation}
	where $\Delta^{\pi_{\mathrm{old}}}$ is the partition function that normalizes the distribution.
		
	\subsection{Distributional Reinforcement Learning}
		
	Distributional RL extends MaxEnt RL by accounting for the randomness in accumulated discounted returns. The objective function of distributional RL aligns with traditional RL, aiming to maximize the discounted return as in Equation~\ref{equ:rl-obj}.
		
	Let $\mathcal{Z}$ denote the action-value distribution space with finite moments. The \emph{distributional Bellman operator} $\mathcal{T}_{D}^\pi: \mathcal{Z}\to\mathcal{Z}$~\cite{c51} is defined as:
	\begin{equation}
		\label{equ:dist-op}
		\mathcal{T}_{D}^\pi Z(s, a) :\overset{D}{=} R(s, a) + \gamma Z\left(s',a'\right), s' \sim P(\cdot \mid s, a), a' \sim \pi(\cdot \mid s'),
	\end{equation}
	where $U :\overset{D}{=} V$ indicates that random variables $U$ and $V$ have the same distribution.
		
	In the control setting, the \emph{distributional Bellman optimality operator} $\mathcal {T}_{D}^*$ is defined as:
	\begin{equation}
		\label{equ:dist-op-opt}
		\mathcal{T}_{D}^* Z(s, a) :\overset{D}{=} R(s, a) + \gamma Z\left(s',a^*\right) ,s' \sim P(\cdot \mid s, a), a^* \in \mathop{\arg\max}_{a'} \mathbb{E} [Z(s',a')].
	\end{equation}
		
	To measure the distance between value distributions, the $p$-Wasserstein distance is employed, defined as:
	\begin{equation}
		d_{p}(U, V)=\left(\int_{0}^{1}\left|F_{U}^{-1}(\omega)-F_{V}^{-1}(\omega)\right|^{p} d \omega\right)^{1 / p},
	\end{equation}
	where $F_U$ and $F_V$ are the cumulative distribution functions (CDFs) of two random variables $U$ and $V$. For $Z_1, Z_2 \in \mathcal{Z}$, the supremum-$p$-Wasserstein metric $\bar{d}_p$ over value distributions is given by:
	\begin{equation}
		\bar{d}_p(Z_1, Z_2)=\sup_{s,a} d_{p} \left( Z_1(s,a), Z_2(s,a) \right).
	\end{equation}
	The operator $\mathcal{T}_{D}^\pi$ is a $\gamma$-contraction in $\bar{d}_p$. Although $\mathcal{T}_{D}^*$ does not maintain the contractive property of any distributional metric, convergence to the optimal action-state function $Q^*$ can still be attained by taking the expectation $\mathbb{E} [Z]$~\cite{c51}.
		
	A primary challenge in distributional RL is the approximation of value distributions. Quantile Distributional Reinforcement Learning (QDRL) has emerged as a popular method for distribution approximation~\cite{qr-dqn,iqn,fqf}. In QDRL, the return distribution is projected onto a parameterized quantile distribution, expressed as $Z(s,a)=\frac{1}{K}\sum_{k=0}^{K-1}\delta_{z_k}(s,a)$, where $\delta_z$ is a Dirac measure, $z_k$ is a quantile atom located at the $\tau_k$-quantile with a total number of $K$ atoms. The $d_1$ distance between the quantile distribution and the target is minimized using the quantile regression loss:
	\begin{equation*}
		z_k = \mathop{\arg\min}_q \mathbb{E}_Z \left[ \left( \tau_k \mathbbm{1}_{Z>q} +  (1-\tau_k) \mathbbm{1}_{Z\leq q} \right) |Z-q| \right],
	\end{equation*}
	where $\mathbbm{1}(\cdot)$ is the indicator function and $q$ is the candidate location for the $\tau_k$-quantile..
		
	\section{Distributional Soft Actor-Critic}\label{subsec:DSAC}
	In this section, we introduce our novel algorithm, the \emph{Distributional Soft Actor-Critic} (DSAC). Initially, we propose a new Bellman operator and develop a policy iteration scheme known as Distributional Soft Policy Iteration (DSPI), which is proven to converge to the optimal value distribution. Subsequently, we detail our DSAC algorithm, which effectively implements DSPI within an Actor-Critic framework. The specifics for training the distributional critic using quantile regression and double learning are provided at the end of this section.
		
	\subsection{Distributional Soft Policy Iteration}
		
	DSAC is designed to optimize the policy to maximize the objective function outlined in Equation~\ref{equ:sac-obj}, in line with the principles of maximum entropy reinforcement learning (MaxEnt RL). Accounting for the inherent randomness in both rewards and actions, we define the \emph{soft action-value distribution} $Z^\pi$ for a policy $\pi \in \Pi$ as follows:
	\begin{align}
		\label{equ:def-z}
		Z^\pi(s, a) & := R(s, a) + \sum_{t=1}^\infty \gamma^t \left( R(s_t, a_t) - \alpha \log \pi(a_t \mid s_t) \right), \notag \\
		            & \qquad a_t \sim \pi(\cdot \mid s_t), s_{t+1} \sim P(\cdot \mid s_t, a_t), s_0 = s, a_0 = a.                
	\end{align}
	We introduce the \emph{distributional soft Bellman operator} $\mathcal{T}_{DS}^\pi$ as:
	\begin{align}
		\label{equ:dsac-op}
		\mathcal{T}_{DS}^\pi Z(s, a) & := R(s, a) + \gamma \left( Z\left( s', a' \right) - \alpha \log \pi(a' \mid s') \right), \notag \\
		                             & \qquad s' \sim P(\cdot \mid s, a), a' \sim \pi(\cdot \mid s').                                  
	\end{align}
	This operator harmoniously combines the soft Bellman operator $\mathcal{T}_{S}^\pi$ from Equation~\ref{equ:soft-op} and the distributional Bellman operator $\mathcal{T}_{D}^\pi$ from Equation~\ref{equ:dist-op}, inheriting the convergence property from its constituent operators.
		
	\begin{lemma}
		The operator $\mathcal{T}_{DS}^\pi: \mathcal{Z} \to \mathcal{Z}$ is a $\gamma$-contraction with respect to the distance metric $\bar{d}_p$.
		\label{lem:operator}
	\end{lemma}
		
	Armed with this new operator, we can formulate an algorithm akin to soft policy iteration, termed \emph{distributional soft policy iteration}. This algorithm is divided into two phases: \emph{distributional soft policy evaluation} and \emph{distributional soft policy improvement}. For any given policy $\pi$, the soft action-value distribution $Z^\pi$ is obtained by iteratively applying $\mathcal{T}_{DS}^\pi$.
		
	\begin{lemma}[Distributional Soft Policy Evaluation]
		Let $Z_{k+1} := \mathcal{T}_{DS}^\pi Z_k$ with $Z_0 \in \mathcal{Z}$. The sequence $\{Z_k\}$ will asymptotically converge to $Z^\pi$ as $k \to \infty$.
		\label{lem:evaluation}
	\end{lemma}
		
	Once the value distribution is accurately evaluated, policy improvement is achieved by solving the optimization problem defined in Equation~\ref{equ:sac-ori}.
		
	\begin{lemma}[Distributional Soft Policy Improvement]
		Let $Q^\pi(s,a) := \mathbb{E} [Z^\pi(s,a)]$ for all $(s,a) \in \mathcal{S} \times \mathcal{A}$. Given an old policy $\pi_{\mathrm{old}} \in \Pi$ and a new policy $\pi_{\mathrm{new}}$ obtained by solving the problem in Equation~\ref{equ:sac-ori}, it holds that $Q^{\pi_{\mathrm{old}}}(s,a) \leq Q^{\pi_{\mathrm{new}}}(s,a)$.
		\label{lem:improvement}
	\end{lemma}
		
	The optimal policy $\pi^*$ is derived from $Q^*$, the expectation of the optimal action-value distribution $Z^*$. This derivation involves a many-to-one mapping, as multiple distributions $Z^*$ could yield identical expected values. However, since the entropy regularization term $\alpha \mathcal{H}\left(\pi\left(\cdot \mid s_t \right)\right)$ in DSAC objective function $\mathcal{J}(\theta)$, extended from SAC, is a convex function in $\pi^{*}(\cdot | s)$ at its optimality, this constraint in $\mathcal{J}(\theta)$ allows the optimal policy achieves its uniqueness at minimization for the current policy $\pi_{t}$ \cite{geist2019theory}. Iteratively applying distributional soft policy evaluation and improvement will result in the unique optimal policy $\pi^*$.

	The uniqueness of $\pi^*$ implies the uniqueness of the $Z^*$. This distribution can be obtained by repeatedly applying he distributional soft Bellman operator $\mathcal{T}_{DS}^\pi$ to $\pi_t$. Moreover, to understand the convergence of the soft action-value distribution in the control setting, we define the \emph{distributional soft Bellman 
		optimality operator} $\mathcal{T}_{DS}^*$ as:
	\begin{equation}
		\label{equ:dsac-op-opt}
		\mathcal{T}_{DS}^* = \mathcal{T}_{DS}^\mu, \quad
		\mu(\cdot \mid s) \propto \exp(\mathbb{E} [Z(s,\cdot)]/\alpha).
	\end{equation}
	\begin{theorem}[Convergence in the Control Setting]
		Let $Z_{k+1} := \mathcal{T}_{DS}^* Z_k$ with $Z_0 \in \mathcal{Z}$. The sequence $\{Z_k\}$ will converge to $Z^*$ as $k \to \infty$.
	\end{theorem}
	Unlike $\mathcal{T}_{D}^*$, which leads to a uniform convergence of distributions to a set of optimal state-action distributions~\cite{c51}, the repeated application of $\mathcal{T}_{DS}^*$ results in a unique $Z^*$. This characteristic demonstrates that entropy in the objective function not only enhances exploration but also refines the estimation of the value distribution.
		
	We refer to this phenomenon as the \emph{smoothing effect} of entropy in distribution estimation, which will be further illustrated through a simple experiment in Section~\ref{sec:exp_toy}. Although we have established the convergence of the value distribution, $\mathcal{T}_{DS}^*$ is not a contraction. However, entropy still plays a constructive role in preserving the contractive property, that is, it aids in bringing the updated distribution closer to the optimal distribution. A detailed discussion on this aspect is reserved for the appendix.
		
	The proofs for the aforementioned theoretical results are provided in Appendix~\ref{app:proof}.

	\subsection{The DSAC Algorithm}
		
	The Distributional Soft Actor-Critic (DSAC) algorithm is structured around an Actor-Critic framework, comprising a distributional soft value network, denoted as $Z_\tau(s,a;\theta)$, which serves as the critic, and a stochastic policy network $\pi(a \mid s;\phi)$, which is the actor. We begin by detailing the training process for the critic.
		
	The \emph{quantile function} $F_Z^{-1}$ for a random variable $Z$ is defined as the inverse of its cumulative distribution function (CDF) $F_Z(z) = Pr(Z<z)$. Mathematically, this is expressed as $F_Z^{-1}(\tau) := \inf \{ z \in \mathbb{R} : \tau \leq F_Z(z)\}$, where $\tau$ represents the quantile fraction. In the subsequent discussion, we use $Z_\tau := F_Z^{-1}(\tau)$.
		
	For a given state $s$ and action $a$, the action-value distribution is approximated by a set of quantile fractions $\{\tau_i\}_{i = 0,\dots, N}$, with $\tau_0=0, \tau_N=1$, $\tau_i < \tau_j, \forall i<j$, and $\tau_i \in [0,1], i = 0,\dots, N$. The midpoint of each pair of consecutive quantile fractions is $\hat{\tau}_{i} = (\tau_{i} + \tau_{i+1}) / 2$. Quantile approximation methods include QR-DQN~\cite{qr-dqn}, where the value distribution is approximated by a group of trainable quantile values at fixed quantile fractions. IQN~\cite{iqn} randomly samples the quantile fractions from a uniform distribution. FQF~\cite{fqf} introduces a trainable proposal network to generate $\tau$. Later on, NC-QR-DQN~\cite{nc-qr-dqn} was proposed to solve the quantile crossing issue. NDQFN~\cite{ndqfn} learns a baseline quantile value and then adds non-negative increments to generate monotonic quantile values. A newer method SPL-DQN~\cite{spl-dqn} learns continuous quantile functions represented by monotonic rational quadratic spline. DSAC accommodates these methods flexibly as a modular choice. See appendix~\ref{app:quantile-ablation} for ablation results on QR-DQN, IQN and FQF.

	According to Equation~\ref{equ:dsac-op}, the pairwise temporal difference (TD) error between two quantile fractions $\hat{\tau}_{i}$ and $\hat{\tau}_{j}$ is given by:
	\begin{align}
		\label{equ:dsac-td}                                                                                                                                                                                               
		\delta_{i j}^{t} = r_{t} + \gamma \left[ Z_{\hat{\tau}_{i}} \left(s_{t+1},a_{t+1};\bar{\theta}\right) - \alpha \log \pi(a_{t+1} \mid s_{t+1};\bar{\phi})\right] - Z_{\hat{\tau}_{j}} \left(s_t,a_t;\theta\right), 
	\end{align}
	where $\bar{\theta}$ and $\bar{\phi}$ represent the parameters of the target action-value distribution network and target policy, respectively. The target networks are softly updated to maintain stability during training.
		
	We adopt quantile regression to train the $Z_\tau(s,a;\theta)$ network by minimizing the weighted pairwise Huber regression loss across quantile fractions. This choice is consistent with prior works (QR-DQN, NC-QR-DQN, FQF, NDQFN) to ensure fair comparison across baselines.
	The Huber quantile regression loss~\cite{huber}, with a threshold $\kappa$, is defined as:
	\begin{align*}
		&\varrho_{\tau}^{\kappa}\left(\delta_{i j}\right) =\left|\tau-\mathbbm{1}_{\delta_{i j}<0}\right| \frac{\mathcal{L}_{\kappa}\left(\delta_{i j}\right)}{\kappa}, \text { with } \\
		  & \mathcal{L}_{\kappa}\left(\delta_{i j}\right) =\left\{ \begin{array}{ll}{\frac{1}{2} \delta_{i j}^{2}}, & {\text { if }\left|\delta_{i j}\right| \leq \kappa} \\ 
		{\kappa\left(\left|\delta_{i j}\right|-\frac{1}{2} \kappa\right)}, &{\text { otherwise. }} \end{array} \right.
	\end{align*}
	The objective function for the quantile value network $Z_\tau(s,a;\theta)$ is then:
	\begin{equation}
		\mathcal{J}_Z(\theta) = \sum_{i=0}^{N-1} \sum_{j=0}^{N-1} (\tau_{i+1} - \tau_{i}) \varrho_{\hat{\tau}_{j}}^{\kappa}\left(\delta_{i j}^{t}\right),
	\end{equation}
	where each $\varrho_{\hat{\tau}_j}$ is weighted by the target distribution fractions $\tau_{i+1} - \tau_{i}$.

	\begin{algorithm}[tb]
		\caption{DSAC update}
		\label{alg:dsac}
		\begin{algorithmic}
			\STATE {\bfseries Parameter:} $N, \kappa$
			\STATE {\bfseries Input:} $s, a, r, s',\gamma \in (0,1)$
			\STATE Generate quantile fractions $\tau_i, i=0, \dots, N, \tau_j, j=0, \dots, N$
			\STATE {\color{gray} \# Update Quantile Value Network $Z(s,a; \theta_k)$}
			\STATE Get next actions for calculating target $a'\sim\pi(\cdot \mid s'; \bar{\phi})$
			\FOR{$i=0$ {\bfseries to} $N-1$}
			\FOR{$j=0$ {\bfseries to} $N-1$}
			\STATE $y_i = \min_{k=1,2} Z_{\hat{\tau}_i}\left(s',a';\bar{\theta}_k\right)$
			\STATE $\delta^k_{i j} = r + \gamma \left[y_i   - \alpha \log \pi( a' \mid s';\bar{\phi})\right] - Z_{\hat{\tau}_{j}} \left(s,a;\theta_k\right), k=1,2$
			\ENDFOR
			\ENDFOR
			\STATE $\mathcal{J}_{Z}(\theta_k)=\frac{1}{N} \sum_{i=0}^{N-1} \sum_{j=0}^{N-1} (\tau_{i+1} - \tau_{i}) \varrho_{\hat{\tau}_{j}}^{\kappa}\left(\delta^k_{i j}\right), k=1,2$
			\STATE Update $\theta_k$ with $\nabla \mathcal{J}_{Z}(\theta_k), k=1,2$
			\STATE Update $\bar{\theta}_k \gets \iota \theta_k + (1-\iota) \bar{\theta}_k, k=1,2$ 
			\STATE {\color{gray} \# Update Policy Network $\pi(a \mid s; \phi)$}
			\STATE Get new actions with re-parameterized samples $\tilde{a} \sim \pi(\cdot \mid s; \phi)$
			\STATE $Q(s,\tilde{a})=\sum_{i=0}^{N-1}\left(\tau_{i+1}-\tau_{i}\right) \min_{k=1,2} Z_{\hat{\tau}_i}\left( s,\tilde{a}; \theta_k \right)$
			\STATE $\mathcal{J}_\pi(\phi) = \alpha \log(\pi(\tilde{a} \mid s;\phi)) - Q(s,\tilde{a})$
			\STATE Update $\phi$ with $\nabla \mathcal{J}_\pi(\phi)$
			\STATE Update $\bar{\phi} \gets \iota \phi + (1-\iota) \bar{\phi}$
		\end{algorithmic}
	\end{algorithm}

	DSAC addresses the overestimation issue common in off-policy continuous action RL algorithms when calculating the target value. By extending the double learning concept from TD3 to DSAC, we employ two networks with the same structure, parameterized by $\theta_k, k=1,2$, and trained to fit a conservative target. For two quantile fractions $\hat{\tau}_{i}$ and $\hat{\tau}_{j}$, the TD-error is redefined as:
	\begin{align*}
		y_i^t               & = \min_{k=1,2} Z_{\hat{\tau}_{i}} \left(s_{t+1},a_{t+1};\bar{\theta_k}\right),                                                                     \\
		\delta_{i j}^{t, k} & = r_{t} +\gamma \left( y_i^t - \alpha \log \pi(a_{t+1} \mid s_{t+1};\bar{\phi})\right) - Z_{\hat{\tau}_{j}} \left(s_{t+1},a_{t+1};\theta_k\right). 
	\end{align*}

	The policy network $\pi(a \mid s; \phi)$ is trained by adapting the SAC algorithm using the parameterized quantile function. The action-value function is derived by taking the expectation over the quantile values:
	\begin{equation}
		Q(s, a; \theta)=\sum_{i=0}^{N-1}\left(\tau_{i+1}-\tau_{i}\right) Z_{\hat{\tau}_{i}} \left(s, a;\theta\right).
	\end{equation}
	In the double variant DSAC, $Q(s, a; \theta) = \min_{k=1,2} Q(s,a;\theta_k)$. To solve the minimization problem in Equation~\ref{equ:sac-ori}, SAC samples actions using a re-parameterized policy neural network $f\left(s, \epsilon; \phi\right)$, with $\epsilon$ being a noise vector drawn from a fixed distribution, such as a standard spherical Gaussian. The original problem is then solved using gradient descent with the objective:
	\begin{equation*}
		\mathcal{J}_{\pi}(\phi) = \mathbb{E}_{s_t \sim \mathcal{D}, \epsilon_{t} \sim \mathcal{N}} [\alpha \log \pi\left(f\left(s_t, \epsilon_{t}; \phi\right) \mid s_t \right) - Q\left(s_t, f\left( s_t, \epsilon_{t}; \phi\right); \theta\right) ],
	\end{equation*}
	where $\mathcal{D}$ represents the transitions replay buffer. The complete algorithm is outlined in Algorithm~\ref{alg:dsac}.

	\section{A Risk-sensitive RL Framework}\label{sec:risk}
		
	The distribution of returns encompasses a wealth of information beyond mere expectations, enabling the consideration of diverse risk metrics. In this section, we expand the foundational DSAC to encompass risk-sensitive reinforcement learning (RL) settings, presenting a unified framework capable of optimizing the majority of conventional risk metrics. We designate this risk-sensitive variant of DSAC as RDSAC.
		
	\subsection{Risk-sensitive Policy Learning}
		
	While risk-sensitive RL has been extensively investigated in the literature~\cite{shen2014risk-sensitive,chow2015risk,fu2018risk}, the direct estimation of most risk measures poses a challenge due to the loss of linearity in Bellman's equation~\cite{Puterman1994Markov}. Our approach provides a unified framework that optimizes policies under different risk metrics in complex, continuous control environments. Leveraging the distribution of returns, it approximates the value function under a risk measure.
		
	A \emph{risk measure} $\rho:\mathcal{Z} \to \mathbb{R}$ is a function that maps an uncertain outcome $Z$ to a real number. The expectation is considered a risk-neutral measure function, denoted as $\rho[\cdot]=\mathbb{E}[\cdot]$. Beyond expectation, a multitude of risk preferences can be articulated through risk measures, as discussed in Section~\ref{sec:risk_measures}.
		
	In our approach, we focus on the risk associated with the reward function, rather than the risk of the entropy. To address external rewards and the entropy bonus separately, we partition $Z^\pi$ into two components: the reward distribution $Z^\pi_{R}$ and the entropy distribution $Z^\pi_{H}$:
	\begin{align}
		Z^\pi_{R} & (s, a) :\overset{D}{=} \sum_{t=0}^\infty \gamma^t  R(s_t, a_t), \notag               \\
		Z^\pi_{H} & (s, a) :\overset{D}{=} - \sum_{t=1}^\infty \gamma^t  \log \pi(a_t \mid s_t), \notag  \\
		          & a_t \sim \pi(\cdot \mid s_t), s_{t+1} \sim P(\cdot \mid s_t, a_t), s_0 = s, a_0 = a. 
	\end{align}
	We establish $Z^\pi \overset{D}{=} Z^\pi_R + \alpha Z^\pi_H$, with $Z^\pi_R$ and $Z^\pi_H$ satisfying their respective distributional Bellman equations:
	\begin{align}
		  & \mathcal{T}^\pi_{R} Z(s, a) :\overset{D}{=}  R(s, a) +  \gamma  Z\left( s', a' \right), \notag                                   \\
		  & \mathcal{T}^\pi_{H} Z(s, a) :\overset{D}{=}  \gamma \left( Z\left( s', a' \right) -  \alpha \log \pi(a' \mid s') \right), \notag \\
		  & \qquad s' \sim P(\cdot \mid s, a), a' \sim \pi(\cdot \mid s').                                                                   
	\end{align}
	Consequently, we can train $Z^\pi_R$ and $Z^\pi_H$ using quantile regression. In RDSAC, we employ two quantile networks to parameterize $Z^\pi_R$ and $Z^\pi_H$, sharing parameters except for the final layer of the network.
		
	The risk-sensitive policy is derived by optimizing the following objective:
	\begin{align*}
		\mathcal{J}_{\pi} (\phi) = \mathbb{E}_{s_t \sim \mathcal{D}, a \sim \pi_\phi} [\alpha \log \pi\left(a \mid s_t; \phi \right) -\alpha \mathbb{E} [Z_H\left(s_t, a; \theta_H\right)] - \rho[Z_R \left(s_t, a; \theta_R \right) ]] 
	\end{align*}
	where $\theta_R$ and $\theta_H$ represent the network parameters for $Z^\pi_R$ and $Z^\pi_H$, respectively. By jointly optimizing the entropy and risk value functions, RDSAC integrates risk measures into training while maintaining a diverse policy landscape.
		
	\subsection{Common Risk Measures} \label{sec:risk_measures}
	We detail several prevalent risk measures below, elucidating how RDSAC can optimize them. Initially, we introduce a category of risk measures known as \emph{distorted expectation}, owing to their uniform approximation form. To showcase the adaptability of RDSAC, we also discuss mean-semideviation, a measure that does not fall under the distorted expectation category. MSD penalizes only downside deviations from the mean, making it suitable for risk-averse decision-making scenarios with asymmetrical outcome distributions.
		
	\begin{figure*}[htbp]
		\centering
		\includegraphics[width=0.8\linewidth]{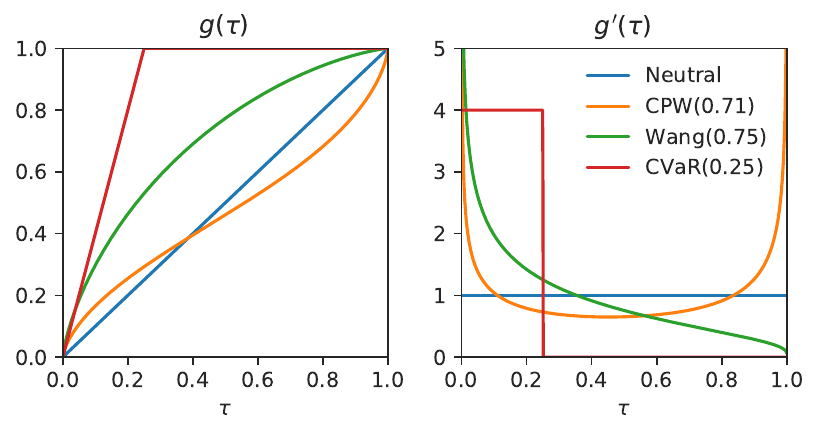}
		\caption{Different distortion functions and their derivations.}
		\Description{Two plots show four distortion functions and their gradients over probability from zero to one. The left plot shows each function’s curvature: CVaR flattens after a threshold, Wang compresses middle probabilities, and CPW emphasizes extremes. The right plot shows gradients: CVaR has a sharp drop, Wang and CPW decay nonlinearly, while Neutral stays constant.}
	\end{figure*}
		
	\paragraph{Distorted Expectation} Distorted expectation is a risk weighted expectation of value distribution under a specific distortion function~\cite{balbas2009properties}. By saying a \emph{distortion function}, we mean a non-decreasing function $g:[0,1] \to [0,1]$ satisfying $g(0)=0$ and $g(1)=1$. The distorted expectation of $Z$ under $g$ is defined as $\rho_{\rm DE}[Z] = \int_0^1 F_Z^{-1}(\tau) d g(\tau)$.
		
	Different distortion functions encapsulate distinct risk propensities. We enumerate some prevalent distortion functions as follows:
		
	\begin{itemize}
		\item \textbf{CVaR} The \emph{Conditional Value at Risk} (CVaR) quantifies risk as the conditional expectation of losses exceeding (rewards below) a specified quantile $\beta$~\cite{rockafellar2000optimization,chow2015risk}. For a random variable $Z$ with the CDF $F_Z(z)$ and a confidence level $\beta \in [0,1]$, CVaR is expressed as $\rho_{\rm CVaR}[Z] = \mathbb{E} \left[ Z \mid Z \leq F_Z^{-1}(\beta) \right]$. The corresponding distortion function is defined as $g(\tau) = \min \{ \tau / \beta, 1 \}$. Estimating CVaR precisely is complex, as only $\beta$ fraction of the data can influence the CVaR value.
		      		      
		\item \textbf{Wang} The \textit{Wang} distortion risk measure~\cite{wang} is given by $g(\tau) = \Phi(\Phi^{-1}(\tau)+\beta)$, where $\Phi$ and $\Phi^{-1}$ represent the standard Normal CDF and its inverse, respectively. $\beta>0$ indicates risk aversion, while $\beta<0$ suggests risk seeking. Unlike CVaR, Wang distributes weights across the entire $\tau$ interval and modulates the weight intensity based on $\beta$.
		      		      
		\item \textbf{CPW} \textit{Cumulative Probability Weighting} (CPW) parameterization is defined as $g(\tau) = \tau^{\beta}/\left(\tau^{\beta}+(1-\tau)^{\beta}\right)^{\frac{1}{\beta}}$. It originates from cumulative prospect theory~\cite{cpt}. CPW is sensitive to the extremities when $\tau \to 0$ or $\tau \to 1$, mirroring human decision-makers.~\citet{cpt} determined that $\beta = 0.71$ closely aligns with human subjects. CPW is not strictly risk-seeking or risk-averse but represents a hybrid.
	\end{itemize}
		
	We propose two methods for approximating distorted expectation with sampling. First, drawing inspiration from inverse transform sampling, we have $\rho_{\rm DE}[Z] = \int_0^1 F_Z^{-1}(g^{-1}(\iota)) d \iota$, where $\iota = g(\tau)$. Thus, we can estimate the distorted action-value as:
	\begin{equation*}
		\hat{\rho}_{\rm DE}[Z(s, a)]=\sum_{i=0}^{N-1}\left(\tau_{i+1}-\tau_{i}\right) Z_{g^{-1}(\hat{\tau}_{i})}(s,a; \theta).
	\end{equation*}
	This method is viable when $g^{-1}$ has a closed-form solution, such as with CVaR and Wang. Second, recognizing that $\rho_{\rm DE}[Z] = \int_0^1 g'(\tau)F_Z^{-1}(\tau) d \tau$, it is evident that the expectation is distorted by $g'(\tau)$. Consequently, the distorted action-value can be approximated as:
	\begin{equation*}
		\hat{\rho}_{\rm DE}[Z(s, a)]=\sum_{i=0}^{N-1}(\tau_{i+1}-\tau_{i})g'(\hat{\tau}_{i}) Z_{\hat{\tau}_{i}}(s,a; \theta).
	\end{equation*}
	This approach is primarily utilized for CPW.

	\paragraph{Mean-Semideviation} Variance-related metrics form another significant class of risk measures~\cite{sobel1982variance,castro2012policy,prashanth2016variance-constrained}. While variance is an intuitive measure of uncertainty, fluctuation, and robustness by incorporating higher moment information, semideviation~\cite{ogryczak1999stochastic} distinguishes itself by capturing the variation on only one side of the return distribution. Specifically, the square root of semivariance~\cite{markowitz1959portfolio,Ma2022MeanSemivariancePO} is known as the semideviation (SD). For risk-averse policies, we penalize the downside semideviation as $\mathbb{SD}[Z] := \mathbb{E} [(Z - \mathbb{E} [Z])^2_-]^{1/2}$, where $(\cdot)_- = \min(\cdot, 0)$. Rather than optimizing SD directly, we often consider a mean-semidiviation (MSD) objective $\rho_{\rm MSD}[Z] = \mathbb{E}[Z] - \beta \mathbb{SD}[Z]$, which balances the mean with semidiviation.
		
	As pointed out by \citet{c51}, the distributional Bellman operator $\mathcal{T}_{D}^{\pi}$ is a contraction in variance, which means that convergence in the value distribution space also leads to a good estimation of variance. Under the distributional Bellman operator $\mathcal{T}_{D}^{\pi}$, the variance can be approximated as:
		$$\hat{\rho}_{\mathrm{SD}}[Z(s,a)] =\sqrt{\sum_{i=0}^{N-1}\left(\tau_{i+1}-\tau_i\right)\left[Z_{\tau_i}(s, a ; \theta)-Q(s, a)\right]_{-}^2}$$
		the MSD can be approximated as: 
		$$
		\hat{\rho}_{\mathrm{MSD}}[Z(s,a)] = Q(s,a)-\beta \sqrt{\sum_{i=0}^{N-1}\left(\tau_{i+1}-\tau_i\right)\left[Z_{\tau_i}(s, a ; \theta)-Q(s, a)\right]_{-}^2}
		$$
		where $Q(s,a)=\mathbb{E}[Z(s,a)]$ is the expected value of the return distribution.

	\section{Experiments} 
	\label{sec:experiments}
	In this section, we conduct experiments to answer the following questions:
	\begin{itemize}
		\item Are MaxEnt RL and distributional RL better together than they are alone?
		\item Why would incorporating the policy entropy into distribution learning be helpful?
		\item Can the agent learn in high-stack, complex environments?
		\item How the policy performs with different risk measure?
	\end{itemize}
	Three groups of experiments are designed to address these questions above.

	\subsection{Toy Example} \label{sec:exp_toy}
		
\begin{figure}[htbp]
		\centering
		\includegraphics[width=0.6\columnwidth]{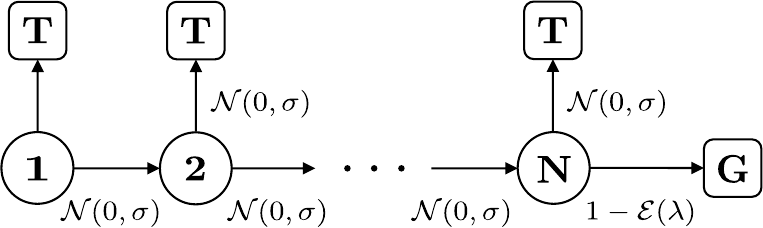}
		\caption{A chain environment.}
		\label{fig:chain_env}
		\Description{The diagram shows a sequential action-state chain graph with nodes labeled 1 to N. Each node transitions to the next via a normal distribution \(\mathcal{N}(0, \sigma)\). Each node also has an upward arrow pointing to a boxed T, representing a transformation. The final node N transitions to a goal node G with probability \(1 - \mathcal{E}(\lambda)\), where \(\mathcal{E}(\lambda)\) is a noise function of lambda parameter.}
	\end{figure}

	We give a toy example to illustrate the benefits of entropy in distributional RL. Here is a chain environment (see Figure~\ref{fig:chain_env}), which contains $N$ states. For each state, the agent can choose either $a_0$ (going right) or $a_1$ (going up). Both actions result in a noise reward $\mathcal{N}(0, \sigma)$, but the episode ends immediately after taking $a_1$. When the agent arrives at $N$-th state and goes right, it will achieve a final noisy reward $1 - \mathcal{E}(\lambda)$. In this paper, we set $\sigma=0.1$ and $\lambda=0.5$. Obviously, the optimal policy is keeping going right.
		
\begin{figure}
		\centering
		    
		\begin{minipage}[b]{\textwidth}
			\centering
			\includegraphics[width=\textwidth]{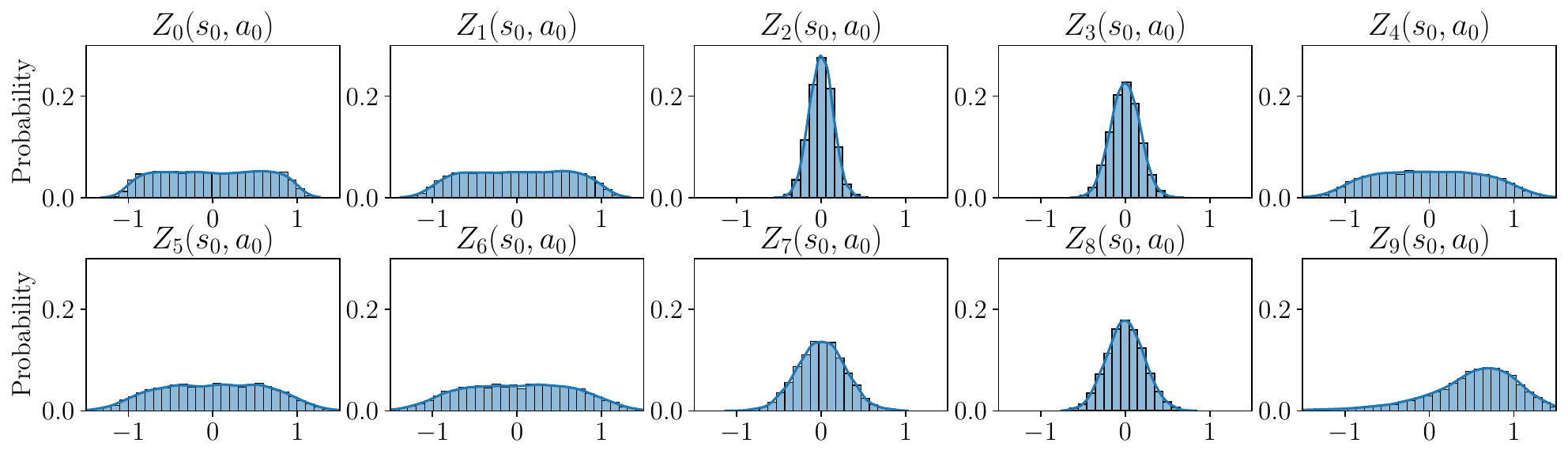}
			{\small (a) $Z_k(s_0, a_0)$ without entropy regularization}
			\label{fig:woent}
                \Description{Histograms of ten target state-action distributions during early training without entropy regularization. Some distributions shift abruptly between steps, showing instability in the learning process.}
		\end{minipage}

		\begin{minipage}[b]{\textwidth}
			\centering
			\includegraphics[width=\textwidth]{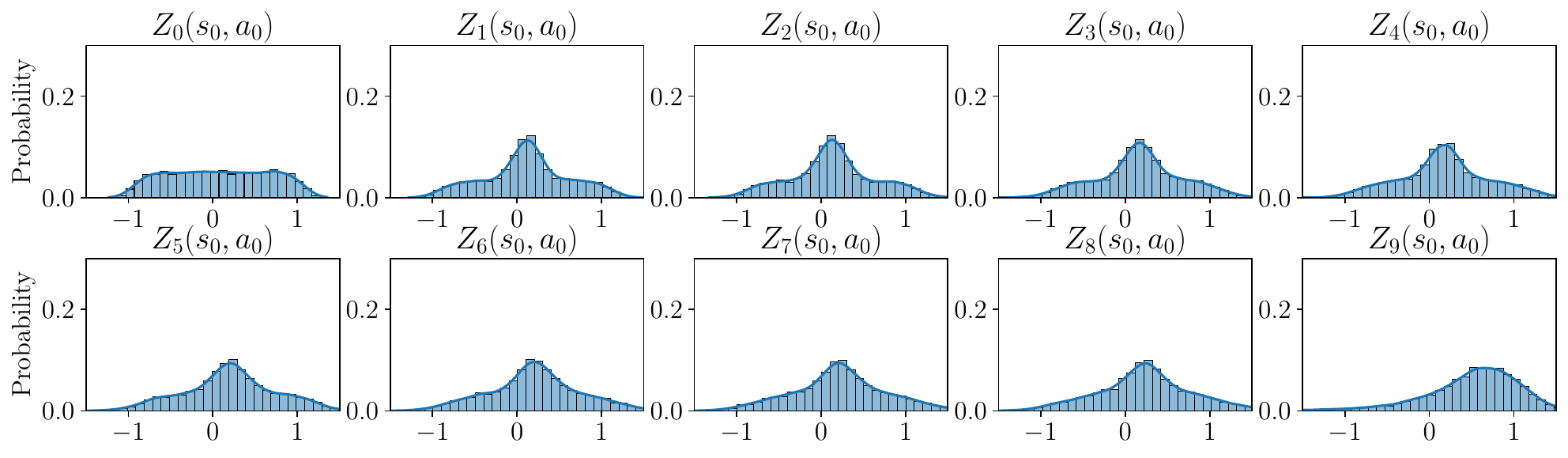}
			{\small (b) $Z_k(s_0,a_0)$ with entropy regularization}
			\label{fig:ent}
                \Description{Histograms of ten target state-action distributions during early training with entropy regularization. Distributions change smoothly between steps, showing stable updates in the learning process.}
		\end{minipage}
		    
		\caption{Visualization of the target state-action distributions $Z_k(s_0,a_0)$ distributions during first 10 iterative updates. $s_0$ is the initial state, the leftmost state in the toy example illustration. $a_0$ is the action of going right (optimal action).}
		\label{fig:demo}
	\end{figure}

	We consider distributional learning with or without entropy regularization respectively. For distributional RL without entropy regularization, the target distribution is taken from the greedy actions directly (see Equation~\ref{equ:dist-op-opt}). Its counterpart, distributional RL with entropy regularization (see Equation~\ref{equ:dsac-op-opt}), targets the distribution as a mixed one under a softmax policy (the optimal policy with entropy regularization). We evaluate the two methods in the chain task above to show the benefit of entropy regularization in distribution learning, and visualize the target distributions of each iteration in Figure~\ref{fig:demo}.
		
	Smoothing involves modifying an objective function or constraints to make them more tractable, particularly with noisy functions. The \emph{smoothing effect} has been observed in the expected RL with entropy on the loss optimization landscape in high dimensional environments~\cite{ahmed2019understanding}. We observes the smoothing effect on target action-value distributions in distributional RL. Since the standard policy learning process only considers the expectation of the whole distribution, the target is sensitive to the noise when Q-values of next actions are similar~\cite{c51}. That phenomenon will do harm to the distributional learning, such as quantile regression. By contrast, the entropy regularized policy is stochastic and stable to the noise in Q estimates, leading to distribution updates with small shifts in training. 
		
	We illustrate such smoothing effect in Figure~\ref{fig:demo} that our target distributions are smoother with the entropy regularization. As demonstrated in toy example, before adding entropy, the distribution $Z_k$ may incur abrupt shifts (Figure~\ref{fig:demo}a). Examples of the abrupt shifts include from $Z_1$ to $Z_2$, from $Z_3$ to $Z_4$ and from $Z_6$ to $Z_7$. After adding entropy in distributional RL, the learning distribution $Z_k$ exhibited small shifts from the previous steps (Figure~\ref{fig:demo}b). We see that adding entropy leads to faster convergence.
	    	
	\subsection{Comparison with baselines}
		
	To evaluate our algorithm performance, we design a series of experiments to compare DSAC with some baselines for continuous control RL algorithms and test DSAC in different risk scenarios. We implement our algorithm based on \emph{rlpyt}~\cite{rlpyt}, a well-developed PyTorch~\cite{pytorch} RL toolkit. All experiments are performed on a servers with 2 AMD EPYC 7702 64-Core Processor CPUs, 2x24-core Intel(R) Xeon(R) Platinum 8268 CPUs, and 8 Nvidia GeForce RTX 2080 Ti GPUs. Hyper-parameters and implementation details are listed in Appendix~\ref{app:hyper-param}. The source code of our DSAC implementation\footnote[1]{https://github.com/xtma/dsac} is available online.

	\begin{figure*}[ht]
		\begin{center}
			\centerline{\includegraphics[width=\linewidth]{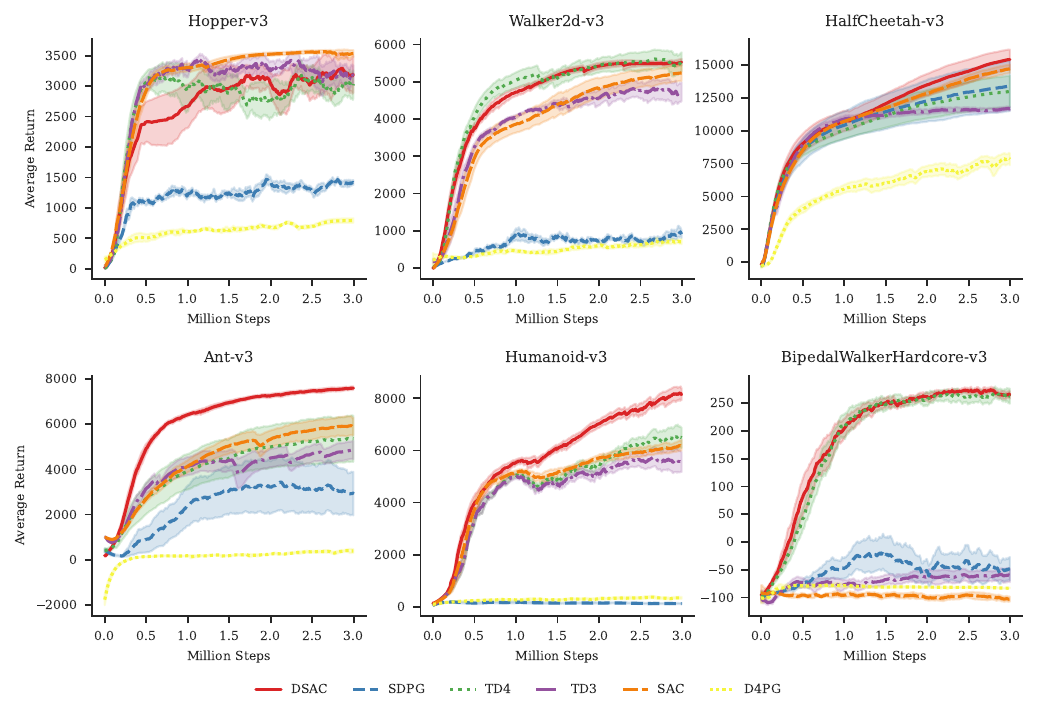}}
			\caption{Learning curves for the continuous control benchmarks in MuJoCo and Box2d. Each learning curve is averaged over 6 different random seeds and shaded by the half of their variance. All curves are smoothed for better visibility.}
			\label{fig:baselines}
                \Description{The line plot shows average return over three million training steps for six environments comparing six reinforcement learning algorithms. The vertical axis shows average return and the horizontal axis shows training progress in millions of steps. DSAC shows faster improvement and higher final performance in most environments. SAC and TD3 also perform well but with more variance. SDPG and D4PG underperform consistently.}
		\end{center}
	\end{figure*}

	We evaluate our algorithm with MuJoCo~\cite{mujoco} and Box2d in OpenAI Gym~\cite{gym}. We compare with two baselines for continuous control tasks: Twin Delayed Deep Deterministic policy gradient algorithm (TD3)~\cite{td3} and Soft Actor-Critic (SAC)~\cite{sac}. To test whether the policy entropy is effective in distributional RL, we compare DSAC with two distributional RL methods, Sample-based Distributional policy gradient (SDPG)~\cite{sdpg} and Distributed Distributional Deep Deterministic Policy Gradient algorithm (D4PG)~\cite{d4pg}, both distributional extension of DDPG. We also implement a distributional variant TD3 named Twin Delayed Deep Distributional Deterministic policy gradient (TD4), which includes techniques such as double learning to reduce the function approximation error. D4PG, SDPG, and TD4 adopt the same distributional critic as DSAC while excluding entropy information in training.

	\begin{table}[htbp]
		\caption{Comparison of average maximal returns ± one standard deviation of 3 million training steps over 6 random seeds. The evaluations are performed every 5000 training steps in each trial over 10 episodes. ``BWH'' denotes BipedalWalkerHardcore.}
		\centering
		\begin{small}
			\begin{tabular}{lrrrrrrr}
				\toprule
				\textbf{Environment} & \textbf{DSAC}         & \textbf{TD4}        & \textbf{SDPG}    & \textbf{SAC}    & \textbf{TD3}               & \textbf{D4PG}             \\
				\midrule
				Hopper-v3            & $\bm{4138 \pm 113}$   & $4034 \pm 131$      & $3712 \pm 234$   & $3634 \pm 60$   & $3814 \pm 83$   & $1543 \pm 451$ \\
				Walker2d-v3          & $5624 \pm 182$        & $\bm{5797 \pm 536}$ & $4459 \pm 956$   & $5361 \pm 442$  & $5364 \pm 565$  & $2016 \pm 346$ \\
				HalfCheetah-v3       & $\bm{15725 \pm 1453}$ & $13396 \pm 2510$    & $13865 \pm 3731$ & $15148 \pm 632$ & $12149 \pm 419$ & $9715 \pm 763$ \\
				Ant-v3               & $\bm{7729 \pm 142}$   & $5791 \pm 2155$     & $5597 \pm 1820$  & $6387 \pm 513$  & $5224 \pm 811$  & $1732 \pm 502$ \\
				Humanoid-v3          & $\bm{8673 \pm 460}$   & $7256 \pm 961$      & $475 \pm 73$     & $6547 \pm 507$  & $6119 \pm 520$  & $641 \pm 174$  \\
				BWH-v3               & $\bm{318 \pm 3}$      & $318 \pm 7$         & $118 \pm 124$    & $-36 \pm 11$    & $-13 \pm 16$    & $5 \pm 70$    \\
				\bottomrule
			\end{tabular}
		\end{small}
		\label{tab:results}
	\end{table}
		
	The results in Figure~\ref{fig:baselines} and Table~\ref{tab:results} show that DSAC outperforms other baselines. Moreover, in complex tasks such as Humanoid-v2 (which has a 17-dimensional action space) and BipealWalkerHardcore-v3 (hard for exploration), DSAC has significant advantages against other methods. Although TD4 also achieves sound results in many tasks like Walker2d-v3, DSAC has better performances in challenging environments with larger action spaces, which demonstrates the entropy regularized objective leads to better exploration.
	It needs to be pointed out that, unlike former solutions combining evolution strategy, implementing recurrent policy or using specific tricks, DSAC finishes BipealWalkerHardcore-v3 with an ideal score without any targeted changes. The results illustrate the effectiveness of modeling distributional information of both reward and action.
		
\begin{figure}[htbp]
		\centering
		\includegraphics[width=0.7\columnwidth]{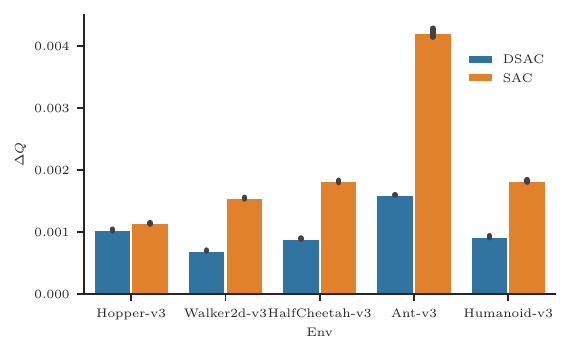}
		\caption{Compare $\Delta Q$ of DSAC and SAC over the last million training steps in MuJoCo tasks.}
		\label{fig:delta_q}
            \Description{Bar chart comparing the change in estimated action value, delta Q, for DSAC and SAC across five MuJoCo environments. The vertical axis shows delta Q values and the horizontal axis lists environments. In all environments, SAC has higher delta Q than DSAC, with the largest gap observed in Ant-v3. This indicates more stable value estimation in DSAC during the final phase of training.}
	\end{figure}

	To help understand the effectiveness of distributional RL, we compare the difference of $Q$-values estimated by DSAC and SAC. As both algorithms utilize double learning for value evaluation, it is a natural way to compare the differences in the output values of the two networks. To be specific, we calculate normalized difference $\Delta Q = 2|Q_1 - Q_2| / |Q_1 + Q_2|$ over the last million steps. The results in Figure~\ref{fig:delta_q} show that DSAC has lower difference between the two network outputs. Reasonably, DSAC trains two $Z$ networks, then takes the smaller of the two and then takes the expected value to get $Q$ values:
	
		\[
			Q_{DSAC}(s,a) =\mathbb{E}[\min(Z_{\tau}(s, a; \theta_1), Z_{\tau}(s, a; \theta_2))].
		\]

		SAC, on the other hand, picks the min of the $Q$ values:
		\[
			Q_{SAC}(s,a) =\min(Q(s, a;\theta_1), Q(s, a;\theta_2))
		\]
	
	where \( Q(\cdot;\theta) = \mathbb{E}[Z(\cdot;\theta)] \).
		As we see $Q_{DSAC}(s,a) \leqslant Q_{SAC}(s,a)$, DSAC algorithm produces smaller estimates, thereby more accurate value networks.

	\subsection{Risk-sensitive Policies}

	The standard MuJoCo environments are deterministic, which limits their applicability in scenarios involving risk. To assess the impact of risk-sensitive measures on agent performance, we utilize a set of tasks termed \emph{risky robot navigation}~\cite{Ma2021ConservativeOD}.

	\paragraph{Risky Mass Point} This task involves a mass point that must navigate to a target goal while avoiding a danger zone. Penalties are incurred upon entry, with the likelihood and severity increasing as the mass point approaches the center of the zone. In our experiments, the danger zone is a circular area with a radius of $d_0=0.3$, centered at $(0.5, 0.5)$. The risk of penalty is modeled by the function $p=p_0e^{-4(d/d_0)^2}$, where $p_0=0.1$ represents the baseline risk and $d$ is the distance to the zone's center. The penalty for entering the danger zone is set at $10$. The mass point is considered to have reached the goal if it is within a circle centered at $(0, 0)$ with a radius of $0.05$. The reward function is designed to encourage proximity to the goal and swift task completion, inversely related to the distance to the goal and reduced by a constant factor. Initial states are randomly selected from $U(0.3, 1)$ for both $x$ and $y$ coordinates, excluding the danger zone.

	\begin{table}[htbp]
		\caption{Comparison of the final performance of 200 thousand training steps over 5 random seeds in Risk Mass Point. The evaluation of each trial is over 100 episodes.}
		\centering
		\begin{small}
			\begin{tabular}{lrrrrr}
				\toprule
				\textbf{Metric} & \textbf{Average}      & \textbf{CVaR(0.25)}   & \textbf{CVaR(0.1)}     & \textbf{Min}           \\
				\midrule
				Neutral         & $-7.20 \pm 0.14$      & $-10.53 \pm 0.44$     & $-13.61 \pm 0.91$      & $-19.25 \pm 0.59$      \\
				CVaR(0.25)      & $-7.13 \pm 0.08$      & \bm{$-9.59 \pm 0.21$} & $-11.04 \pm 0.41$      & $-16.34 \pm 0.49$      \\
				CVaR(0.1)       & $-7.35 \pm 0.09$      & $-9.76 \pm 0.15$      & \bm{$-10.80 \pm 0.20$} & \bm{$-14.17 \pm 0.80$} \\
				Wang(0.75)      & \bm{$-7.12 \pm 0.09$} & $-9.67 \pm 0.27$      & $-11.28 \pm 0.51$      & $-16.95 \pm 1.24$      \\
				CPW(0.71)       & $-7.16 \pm 0.11$      & $-10.21 \pm 0.30$     & $-12.89 \pm 0.61$      & $-19.11 \pm 1.30$      \\
				MSD(1)          & $-7.20 \pm 0.10$      & $-9.66 \pm 0.20$      & $-11.01 \pm 0.34$      & $-15.33 \pm 0.98$      \\
				\bottomrule
			\end{tabular}
		\end{small}
		\label{tab:results_point}
	\end{table}
	    
	\begin{figure}[htbp]
		\centering
		        
		\begin{minipage}[b]{0.32\linewidth}
			\centering
			\includegraphics[width=\linewidth]{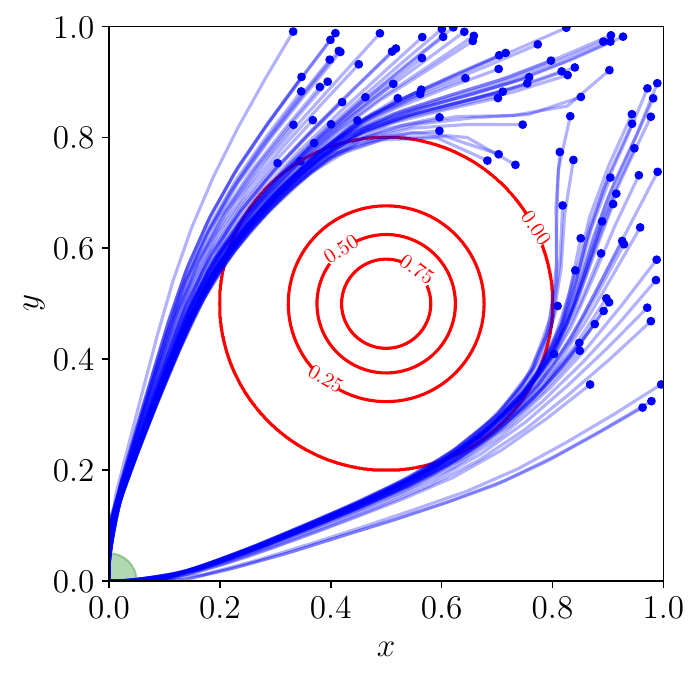}
			{\small (a) CVaR(0.1)}
		\end{minipage}
		\hspace{0.01\linewidth}
		\begin{minipage}[b]{0.32\linewidth}
			\centering
			\includegraphics[width=\linewidth]{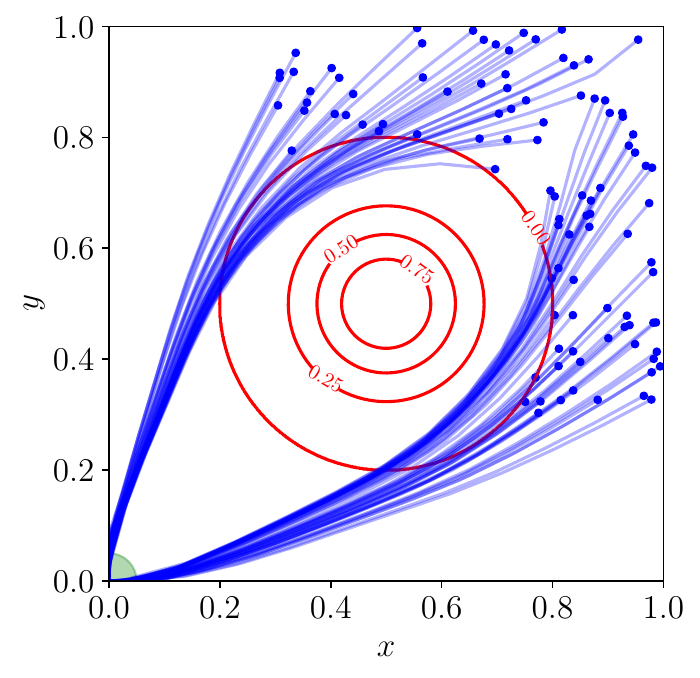}
			{\small (b) CVaR(0.25)}
		\end{minipage}
		\hspace{0.01\linewidth}
		\begin{minipage}[b]{0.32\linewidth}
			\centering
			\includegraphics[width=\linewidth]{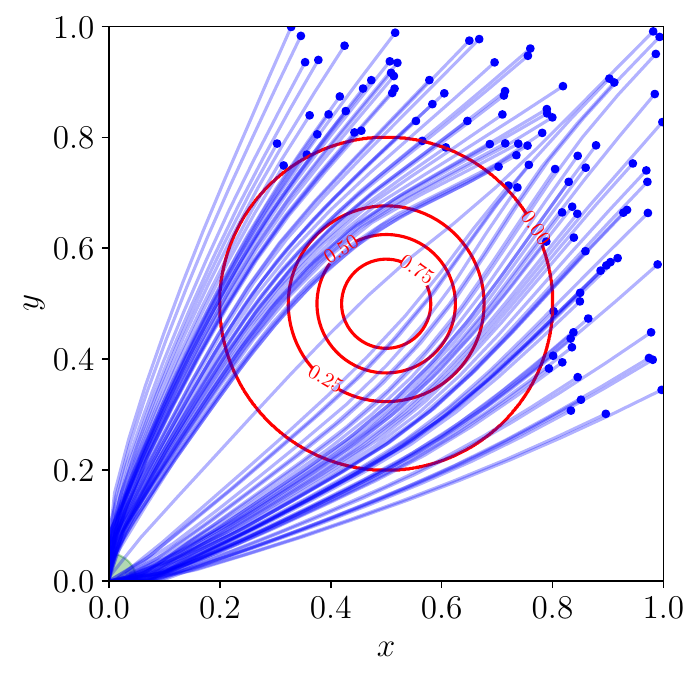}
			{\small (c) Neutral}
		\end{minipage}
		        
		\caption{Visualization of sampled trajectories in Risk Mass Point. For each risk measure, we choose its best seed in terms of average return. The results of other risk measures can be found in Figure~\ref{fig:risk_point_add} in Appendix.}
		\label{fig:risk_point}
            \Description{Three contour plots show two-dimensional trajectory distributions with red distortion function level sets and blue sample paths. The x-axis and y-axis range from zero to one, representing normalized space. Each plot shows multiple blue trajectories originating from the lower left corner, with end points marked as blue dots. The red concentric contours represent distortion function values from 0.25 to 1.00. In the first and second plots, most trajectories follow a curved path along the left boundary, concentrating near the top right. In the third plot, trajectories spread more broadly across the space, showing less alignment. The overlaid distortion contours help visualize how the trajectories interact with different weighting levels in distorted expectation space.}
	\end{figure}

	\paragraph{Risky Ant} To evaluate the algorithm's capability in complex navigation tasks requiring control, we introduce an Ant agent in a larger environment. The Ant replaces the mass point, and the task retains the core elements of the Risky Mass Point. The target area is a circular region with a radius of $0.5$ on the outer edge of a $10 \times 10$ playground. The danger zone, centered at $(5,5)$, has a radius of $d_0=3$. The risk of penalty and its calculation remain consistent with the previous task, but the penalty is increased to $200$. The reward function now includes a velocity bonus to incentivize rapid task completion, shifting from a constant factor. Initial states are sampled from $U(0, 7)$.
		
	For each task, we execute the RDSAC algorithm with various risk measures, selecting one or two representative parameters for each. We assess the learned policies over 100 episodes, calculating the average return, CVaR(0.25), CVaR(0.1), and minimum values to compare their final performances. The results are detailed in Table~\ref{tab:results_point} and Table~\ref{tab:results_ant}.

	In the Risky Mass Point experiments, all risk measure policies successfully navigate the task with comparable average returns. Notably, the Wang(0.75) measure demonstrates superior average performance. Both CVaR(0.1) and CVaR(0.25) excel under their respective metrics, with CVaR(0.1) also achieving the highest minimum value. This is expected, as these measures align with the task's objectives. To further elucidate their behavior, we present selected trajectories in Figure~\ref{fig:risk_point}.

	\begin{table}[htbp]
		\caption{Comparison of the final performance of 3 million training steps over 5 random seeds in Risky Ant. The evaluation of each trial is over 100 episodes.}
		\centering
		\begin{small}
			\begin{tabular}{lrrrrr}
				\toprule
				\textbf{Metric} & \textbf{Average}         & \textbf{CVaR(0.25)}      & \textbf{CVaR(0.1)}       & \textbf{Min}              \\
				\midrule
				Neutral         & $-441.52 \pm 33.79$      & $-671.20 \pm 75.88$      & $-812.97 \pm 111.65$     & $-1409.28 \pm 363.10$     \\
				CVaR(0.25)      & $-427.09 \pm 8.62$       & $-646.40 \pm 16.44$      & $-775.53 \pm 24.83$      & $-1122.70 \pm 89.03$      \\
				CVaR(0.1)       & $-920.45 \pm 375.10$     & $-1370.94 \pm 589.30$    & $-1622.94 \pm 673.30$    & $-2087.99 \pm 644.82$     \\
				Wang(0.75)      & \bm{$-383.50 \pm 10.52$} & \bm{$-554.48 \pm 23.01$} & \bm{$-650.78 \pm 35.91$} & \bm{$-946.51 \pm 157.67$} \\
				CPW(0.71)       & $-423.44 \pm 19.23$      & $-644.12 \pm 27.95$      & $-778.95 \pm 27.19$      & $-1244.84 \pm 99.60$      \\
				MSD(1)          & $-396.79 \pm 10.38$      & $-590.98 \pm 19.72$      & $-717.52 \pm 30.66$      & $-1276.15 \pm 33.27$      \\
				\bottomrule
			\end{tabular}
		\end{small}
		\label{tab:results_ant}
	\end{table}

	\begin{figure}[htpb]
		\centering
		        
		\begin{minipage}[b]{0.32\linewidth}
			\centering
			\includegraphics[width=\linewidth]{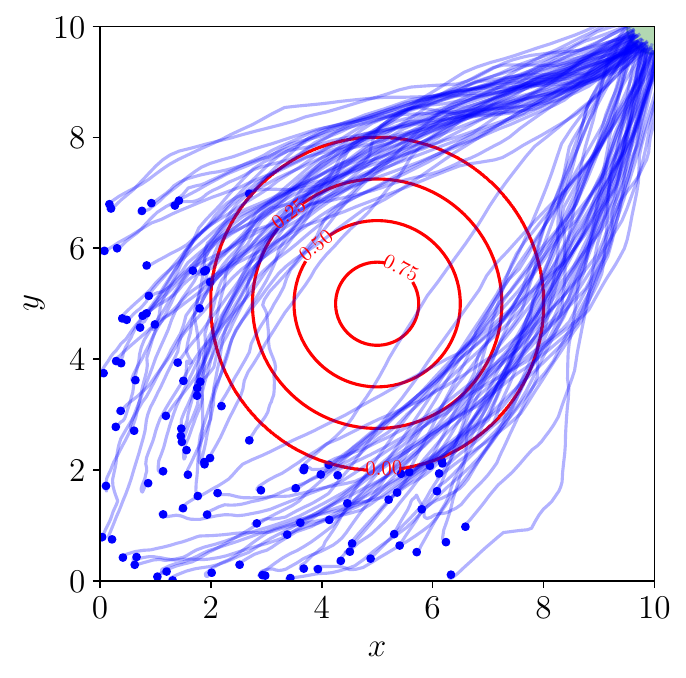}
			{\small (a) CVaR(0.1)}
		\end{minipage}
		\hspace{0.01\linewidth}
		\begin{minipage}[b]{0.32\linewidth}
			\centering
			\includegraphics[width=\linewidth]{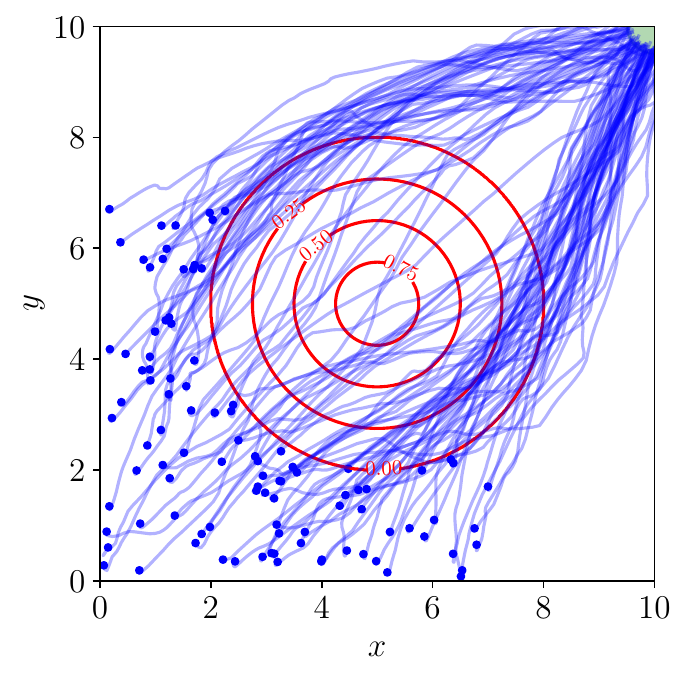}
			{\small (b) CVaR(0.25)}
		\end{minipage}
		\hspace{0.01\linewidth}
		\begin{minipage}[b]{0.32\linewidth}
			\centering
			\includegraphics[width=\linewidth]{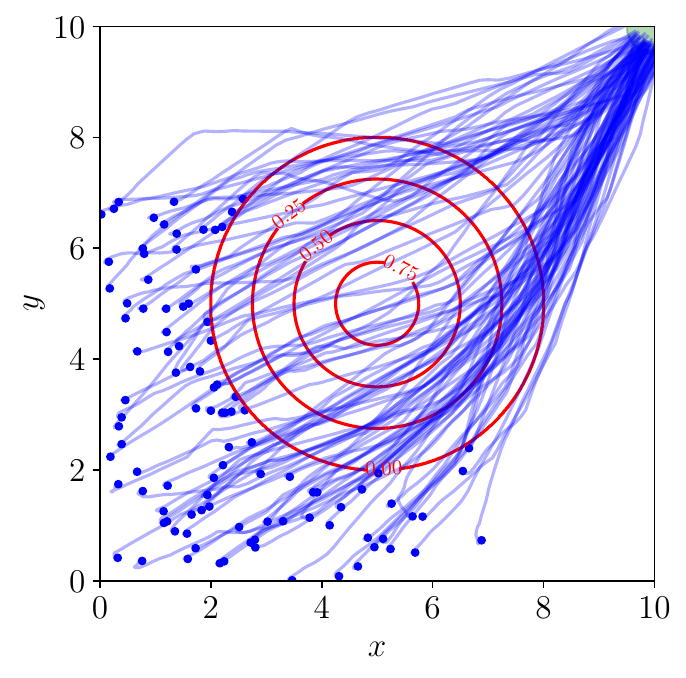}
			{\small (c) Neutral}
		\end{minipage}
		        
		\caption{Visualization of sampled trajectories in Risk Ant. For each risk measure, we choose its best seed in terms of average return. The results of other risk measures can be found in Figure~\ref{fig:risk_ant_add} in Appendix.}
		\label{fig:risk_ant}
            \Description{Three contour plots show two-dimensional trajectories in a 10 by 10 space with red distortion level sets and blue sample paths. Each plot contains trajectories starting from the lower left and ending near the upper right, with end points marked as blue dots. Red contours represent distortion function levels from 0.25 to 1.00. In the first plot, more trajectories pass through high-risk regions indicated by the outer red contours. The second plot shows slightly improved avoidance of these regions. The third plot shows most trajectories staying outside the high-risk zones, indicating improved risk-aware planning.}
	\end{figure}

	Risky Mass Point is a task with simple dynamics, while Risky Ant is more complicated as the agent needs to learn control policy and navigation together. As a result, the overly conservative measure like CVaR(0.1) fails in finishing the navigation task within the training time. A major reason is that a risk-averse objective hinders the exploration since CVaR(0.1) puts too little weight on the part of distribution with larger rewards. In comparison, CVaR(0.25) is less conservative than CVaR(0.1) and accomplished the task while maintaining good risk aversion in training. The sampled trajectories of Risky Ant are shown in Figure~\ref{fig:risk_ant}.
		
	Wang(0.75) balances the return and risk more softly and achieves the best performance in all the metrics. It implies that a softer risk measure might be preferable for complex control tasks to ensure sufficient exploration in training. Meanwhile, MSD(1) also performs well in terms of both return and risk metrics and may be worth trying in other tasks. CPW(0.71) is better than Neutral, but worse than all other measures. Therefore, based on this task, Wang and MSD are recommended as risk measures.

	\section{Conclusions, Discussion and Future Works}
		
	Our proposed algorithm, DSAC (Distributional Soft Actor-Critic), merges MaxEnt RL and distributional RL, considering the randomness present in actions and discounted returns. Under this integration, we introduce a Bellman operator that retains the contractive properties characteristic of MaxEnt RL and at the same time incorporates the flexibility of distributional RL. DSAC surpasses existing baselines across a few continuous control benchmarks. Furthermore, we extended DSAC into a framework for handling diverse risk measures. Our framework Risk-Sensitive Distributional Soft Actor-Critic (RDSAC) balances risk and reward through multiple risk measures in two risk-averse tasks. 
    
    One future work to further show the effectiveness of our unified framework is to include a comparison study with prior works that can also handle multiple risk measures~\cite{iqn,singh2020improving}. \revision{A natural extension to our work is to replace the quantile regression loss in DSAC with a classification-based objective using cross-entropy over discrete bins, which has been shown to yield more expressive representations of value function and provide robustness against noisy target values. ~\cite{farebrother2024revisiting}. Another promising direction is to extend DSAC and RDSAC to model uncertainty over reward functions, allowing sample-based inference in tasks with implicit or human-derived objectives~\cite{liu2025inferencetimes}. Advancing DRL will require richer analysis of the learned distribution, including their representational capacity, distributional metrics, its interaction with value estimation, and the role of parameterization choices, such as categorical, quantile-based, or mixture models~\cite{Rowland2018AnAnalysisCDRL,Sun2022InterpretingDR, bellemare2019drllinear, ChoiYunho2019DDRL}.}
    
 \revision{In conclusion, distribution contains the full information of a random variable. Considering the distribution of discounted returns in RL can help treat various optimization objectives, such as expected objective or risk objectives. How to develop efficient optimization algorithms for distributional RL is a promising research direction. The DSAC and RDSAC algorithms in this paper offer a robust foundation for future research and practical applications, particularly in domains where the trade-off between risk and reward is critical.}

	\begin{acks}
		\revision{This work was supported in part by the National Key Research and Development Program of China (2022YFA1004600), the National Natural Science Foundation of China (72342006, 72371253, 72461160315), the Guangdong Basic and Applied Basic Research Foundation (2023A1515012492, 2023B1515040001), the Regional Joint Foundation of Guangdong (2022A1515110725), and the Guangdong Province Key Laboratory of Computational Science at the Sun Yat-sen University. The work described in this paper was also partially supported by InnoHK initiative, the Government of the HKSAR, and Laboratory for AI-Powered Financial Technologies. This work was also generously supported by an NSF grant ECCS 2419564.}

	\end{acks}
	\bibliographystyle{ACM-Reference-Format}
	\bibliography{citation}


\begin{thebibliography}{72}

%%% the Web address.

\ifx \showCODEN    \undefined \def \showCODEN     #1{\unskip}     \fi
\ifx \showDOI      \undefined \def \showDOI       #1{#1}\fi
\ifx \showISBNx    \undefined \def \showISBNx     #1{\unskip}     \fi
\ifx \showISBNxiii \undefined \def \showISBNxiii  #1{\unskip}     \fi
\ifx \showISSN     \undefined \def \showISSN      #1{\unskip}     \fi
\ifx \showLCCN     \undefined \def \showLCCN      #1{\unskip}     \fi
\ifx \shownote     \undefined \def \shownote      #1{#1}          \fi
\ifx \showarticletitle \undefined \def \showarticletitle #1{#1}   \fi
\ifx \showURL      \undefined \def \showURL       {\relax}        \fi
% invisible to TeX
\providecommand\bibfield[2]{#2}
\providecommand\bibinfo[2]{#2}
\providecommand\natexlab[1]{#1}
\providecommand\showeprint[2][]{arXiv:#2}

\bibitem[Ahmed et~al\mbox{.}(2019)]%
        {ahmed2019understanding}
\bibfield{author}{\bibinfo{person}{Zafarali Ahmed}, \bibinfo{person}{Nicolas Le~Roux}, \bibinfo{person}{Mohammad Norouzi}, {and} \bibinfo{person}{Dale Schuurmans}.} \bibinfo{year}{2019}\natexlab{}.
\newblock \showarticletitle{Understanding the impact of entropy on policy optimization}. In \bibinfo{booktitle}{\emph{International conference on machine learning}}. PMLR, \bibinfo{pages}{151--160}.
\newblock


\bibitem[Balb{\'a}s et~al\mbox{.}(2009)]%
        {balbas2009properties}
\bibfield{author}{\bibinfo{person}{Alejandro Balb{\'a}s}, \bibinfo{person}{Jos{\'e} Garrido}, {and} \bibinfo{person}{Silvia Mayoral}.} \bibinfo{year}{2009}\natexlab{}.
\newblock \showarticletitle{Properties of distortion risk measures}.
\newblock \bibinfo{journal}{\emph{Methodology and Computing in Applied Probability}} \bibinfo{volume}{11}, \bibinfo{number}{3} (\bibinfo{year}{2009}), \bibinfo{pages}{385}.
\newblock


\bibitem[Barth-Maron et~al\mbox{.}(2018)]%
        {d4pg}
\bibfield{author}{\bibinfo{person}{Gabriel Barth-Maron}, \bibinfo{person}{Matthew~W Hoffman}, \bibinfo{person}{David Budden}, \bibinfo{person}{Will Dabney}, \bibinfo{person}{Dan Horgan}, \bibinfo{person}{Alistair Muldal}, \bibinfo{person}{Nicolas Heess}, {and} \bibinfo{person}{Timothy Lillicrap}.} \bibinfo{year}{2018}\natexlab{}.
\newblock \showarticletitle{Distributed distributional deterministic policy gradients}.
\newblock \bibinfo{journal}{\emph{ArXiv preprint}}  \bibinfo{volume}{abs/1804.08617} (\bibinfo{year}{2018}).
\newblock


\bibitem[Bellemare et~al\mbox{.}(2017)]%
        {c51}
\bibfield{author}{\bibinfo{person}{Marc~G Bellemare}, \bibinfo{person}{Will Dabney}, {and} \bibinfo{person}{R{\'e}mi Munos}.} \bibinfo{year}{2017}\natexlab{}.
\newblock \showarticletitle{A distributional perspective on reinforcement learning}. In \bibinfo{booktitle}{\emph{International Conference on Machine Learning}}. JMLR. org, \bibinfo{pages}{449--458}.
\newblock


\bibitem[Bellemare et~al\mbox{.}(2019a)]%
        {bellemare2019distributional}
\bibfield{author}{\bibinfo{person}{Marc~G Bellemare}, \bibinfo{person}{Nicolas Le~Roux}, \bibinfo{person}{Pablo~Samuel Castro}, {and} \bibinfo{person}{Subhodeep Moitra}.} \bibinfo{year}{2019}\natexlab{a}.
\newblock \showarticletitle{Distributional reinforcement learning with linear function approximation}. In \bibinfo{booktitle}{\emph{The 22nd International Conference on Artificial Intelligence and Statistics}}. \bibinfo{pages}{2203--2211}.
\newblock


\bibitem[Bellemare et~al\mbox{.}(2019b)]%
        {bellemare2019drllinear}
\bibfield{author}{\bibinfo{person}{Marc~G. Bellemare}, \bibinfo{person}{Nicolas~Le Roux}, \bibinfo{person}{Pablo~Samuel Castro}, {and} \bibinfo{person}{Subhodeep Moitra}.} \bibinfo{year}{2019}\natexlab{b}.
\newblock \bibinfo{title}{Distributional reinforcement learning with linear function approximation}.
\newblock
\newblock
\showeprint[arxiv]{1902.03149}~[cs.LG]
\urldef\tempurl%
\url{https://arxiv.org/abs/1902.03149}
\showURL{%
\tempurl}


\bibitem[Bertsekas(1995)]%
        {bertsekas1995dynamic}
\bibfield{author}{\bibinfo{person}{Dimitri~P. Bertsekas}.} \bibinfo{year}{1995}\natexlab{}.
\newblock \bibinfo{booktitle}{\emph{Dynamic Programming and Optimal Control} (\bibinfo{edition}{1st} ed.)}.
\newblock \bibinfo{publisher}{Athena Scientific}.
\newblock
\showISBNx{1886529124}


\bibitem[Bertsekas and Tsitsiklis(1995)]%
        {ndp}
\bibfield{author}{\bibinfo{person}{Dimitri~P. Bertsekas} {and} \bibinfo{person}{John~N. Tsitsiklis}.} \bibinfo{year}{1995}\natexlab{}.
\newblock \showarticletitle{Neuro-dynamic programming: an overview}. In \bibinfo{booktitle}{\emph{Proceedings of 1995 34th IEEE Conference on Decision and Control}}, Vol.~\bibinfo{volume}{1}. IEEE, \bibinfo{pages}{560--564}.
\newblock


\bibitem[Bodnar et~al\mbox{.}(2019)]%
        {bodnar2019quantile}
\bibfield{author}{\bibinfo{person}{Cristian Bodnar}, \bibinfo{person}{Adrian Li}, \bibinfo{person}{Karol Hausman}, \bibinfo{person}{Peter Pastor}, {and} \bibinfo{person}{Mrinal Kalakrishnan}.} \bibinfo{year}{2019}\natexlab{}.
\newblock \showarticletitle{Quantile QT-Opt for Risk-Aware Vision-Based Robotic Grasping}.
\newblock \bibinfo{journal}{\emph{ArXiv preprint}}  \bibinfo{volume}{abs/1910.02787} (\bibinfo{year}{2019}).
\newblock


\bibitem[Brockman et~al\mbox{.}(2018)]%
        {gym}
\bibfield{author}{\bibinfo{person}{Greg Brockman}, \bibinfo{person}{Vicki Cheung}, \bibinfo{person}{Ludwig Pettersson}, \bibinfo{person}{Jonas Schneider}, \bibinfo{person}{John Schulman}, \bibinfo{person}{Jie Tang}, {and} \bibinfo{person}{Wojciech Zaremba}.} \bibinfo{year}{2018}\natexlab{}.
\newblock \showarticletitle{Openai gym}.
\newblock \bibinfo{journal}{\emph{ArXiv preprint}}  \bibinfo{volume}{abs/1606.01540} (\bibinfo{year}{2018}).
\newblock


\bibitem[Cen et~al\mbox{.}(2022)]%
        {cen2020fast}
\bibfield{author}{\bibinfo{person}{Shicong Cen}, \bibinfo{person}{Chen Cheng}, \bibinfo{person}{Yuxin Chen}, \bibinfo{person}{Yuting Wei}, {and} \bibinfo{person}{Yuejie Chi}.} \bibinfo{year}{2022}\natexlab{}.
\newblock \showarticletitle{Fast global convergence of natural policy gradient methods with entropy regularization}.
\newblock \bibinfo{journal}{\emph{Operations Research}} \bibinfo{volume}{70}, \bibinfo{number}{4} (\bibinfo{year}{2022}), \bibinfo{pages}{2563--2578}.
\newblock


\bibitem[Choi et~al\mbox{.}(2019)]%
        {ChoiYunho2019DDRL}
\bibfield{author}{\bibinfo{person}{Yunho Choi}, \bibinfo{person}{Kyungjae Lee}, {and} \bibinfo{person}{Songhwai Oh}.} \bibinfo{year}{2019}\natexlab{}.
\newblock \showarticletitle{Distributional Deep Reinforcement Learning with a Mixture of Gaussians}. In \bibinfo{booktitle}{\emph{Proceedings - IEEE International Conference on Robotics and Automation}}. \bibinfo{publisher}{IEEE}, \bibinfo{pages}{9791--9797}.
\newblock
\showISBNx{9781538660270}
\showISSN{2577-087X}


\bibitem[Chow et~al\mbox{.}(2015)]%
        {chow2015risk}
\bibfield{author}{\bibinfo{person}{Yinlam Chow}, \bibinfo{person}{Aviv Tamar}, \bibinfo{person}{Shie Mannor}, {and} \bibinfo{person}{Marco Pavone}.} \bibinfo{year}{2015}\natexlab{}.
\newblock \showarticletitle{Risk-sensitive and robust decision-making: a {CVaR} optimization approach}. In \bibinfo{booktitle}{\emph{Advances in Neural Information Processing Systems}}. \bibinfo{pages}{1522--1530}.
\newblock


\bibitem[Dabney et~al\mbox{.}(2020)]%
        {dabney2020distributional}
\bibfield{author}{\bibinfo{person}{Will Dabney}, \bibinfo{person}{Zeb Kurth-Nelson}, \bibinfo{person}{Naoshige Uchida}, \bibinfo{person}{Clara~Kwon Starkweather}, \bibinfo{person}{Demis Hassabis}, \bibinfo{person}{R{\'e}mi Munos}, {and} \bibinfo{person}{Matthew Botvinick}.} \bibinfo{year}{2020}\natexlab{}.
\newblock \showarticletitle{A distributional code for value in dopamine-based reinforcement learning}.
\newblock \bibinfo{journal}{\emph{Nature}} \bibinfo{volume}{577}, \bibinfo{number}{7792} (\bibinfo{year}{2020}), \bibinfo{pages}{671--675}.
\newblock


\bibitem[Dabney et~al\mbox{.}(2018a)]%
        {iqn}
\bibfield{author}{\bibinfo{person}{Will Dabney}, \bibinfo{person}{Georg Ostrovski}, \bibinfo{person}{David Silver}, {and} \bibinfo{person}{Remi Munos}.} \bibinfo{year}{2018}\natexlab{a}.
\newblock \showarticletitle{Implicit Quantile Networks for Distributional Reinforcement Learning}. In \bibinfo{booktitle}{\emph{International Conference on Machine Learning}}. \bibinfo{pages}{1104--1113}.
\newblock


\bibitem[Dabney et~al\mbox{.}(2018b)]%
        {qr-dqn}
\bibfield{author}{\bibinfo{person}{Will Dabney}, \bibinfo{person}{Mark Rowland}, \bibinfo{person}{Marc~G Bellemare}, {and} \bibinfo{person}{R{\'e}mi Munos}.} \bibinfo{year}{2018}\natexlab{b}.
\newblock \showarticletitle{Distributional reinforcement learning with quantile regression}. In \bibinfo{booktitle}{\emph{Thirty-Second AAAI Conference on Artificial Intelligence}}.
\newblock


\bibitem[Duan et~al\mbox{.}(2021)]%
        {duan2021distributional}
\bibfield{author}{\bibinfo{person}{Jingliang Duan}, \bibinfo{person}{Yang Guan}, \bibinfo{person}{Shengbo~Eben Li}, \bibinfo{person}{Yangang Ren}, \bibinfo{person}{Qi Sun}, {and} \bibinfo{person}{Bo Cheng}.} \bibinfo{year}{2021}\natexlab{}.
\newblock \showarticletitle{Distributional soft actor-critic: Off-policy reinforcement learning for addressing value estimation errors}.
\newblock \bibinfo{journal}{\emph{IEEE transactions on neural networks and learning systems}} \bibinfo{volume}{33}, \bibinfo{number}{11} (\bibinfo{year}{2021}), \bibinfo{pages}{6584--6598}.
\newblock


\bibitem[Farebrother et~al\mbox{.}(2024)]%
        {farebrother2024revisiting}
\bibfield{author}{\bibinfo{person}{Jesse Farebrother}, \bibinfo{person}{Jordi Orbay}, \bibinfo{person}{Quan Vuong}, \bibinfo{person}{Adrien Ali~Taiga}, \bibinfo{person}{Yevgen Chebotar}, \bibinfo{person}{Ted Xiao}, \bibinfo{person}{Alex Irpan}, \bibinfo{person}{Sergey Levine}, \bibinfo{person}{Pablo~Samuel Castro}, \bibinfo{person}{Aleksandra Faust}, \bibinfo{person}{Aviral Kumar}, {and} \bibinfo{person}{Rishabh Agarwal}.} \bibinfo{year}{2024}\natexlab{}.
\newblock \showarticletitle{Stop Regressing: Training Value Functions via Classification for Scalable Deep {RL}}. In \bibinfo{booktitle}{\emph{Proceedings of the 41st International Conference on Machine Learning}} \emph{(\bibinfo{series}{Proceedings of Machine Learning Research}, Vol.~\bibinfo{volume}{235})}, \bibfield{editor}{\bibinfo{person}{Ruslan Salakhutdinov}, \bibinfo{person}{Zico Kolter}, \bibinfo{person}{Katherine Heller}, \bibinfo{person}{Adrian Weller}, \bibinfo{person}{Nuria Oliver}, \bibinfo{person}{Jonathan Scarlett}, {and} \bibinfo{person}{Felix Berkenkamp}} (Eds.). \bibinfo{publisher}{PMLR}, \bibinfo{pages}{13049--13071}.
\newblock
\urldef\tempurl%
\url{https://proceedings.mlr.press/v235/farebrother24a.html}
\showURL{%
\tempurl}


\bibitem[Fujimoto et~al\mbox{.}(2018)]%
        {td3}
\bibfield{author}{\bibinfo{person}{Scott Fujimoto}, \bibinfo{person}{Herke Hoof}, {and} \bibinfo{person}{David Meger}.} \bibinfo{year}{2018}\natexlab{}.
\newblock \showarticletitle{Addressing Function Approximation Error in Actor-Critic Methods}. In \bibinfo{booktitle}{\emph{International Conference on Machine Learning}}. \bibinfo{pages}{1582--1591}.
\newblock


\bibitem[Geist et~al\mbox{.}(2019)]%
        {geist2019theory}
\bibfield{author}{\bibinfo{person}{Matthieu Geist}, \bibinfo{person}{Bruno Scherrer}, {and} \bibinfo{person}{Olivier Pietquin}.} \bibinfo{year}{2019}\natexlab{}.
\newblock \showarticletitle{A Theory of Regularized Markov Decision Processes}. In \bibinfo{booktitle}{\emph{International Conference on Machine Learning}}. \bibinfo{pages}{2160--2169}.
\newblock


\bibitem[Gu et~al\mbox{.}(2017)]%
        {gu2017deep}
\bibfield{author}{\bibinfo{person}{Shixiang Gu}, \bibinfo{person}{Ethan Holly}, \bibinfo{person}{Timothy Lillicrap}, {and} \bibinfo{person}{Sergey Levine}.} \bibinfo{year}{2017}\natexlab{}.
\newblock \showarticletitle{Deep reinforcement learning for robotic manipulation with asynchronous off-policy updates}. In \bibinfo{booktitle}{\emph{2017 IEEE international conference on robotics and automation (ICRA)}}. IEEE, \bibinfo{pages}{3389--3396}.
\newblock


\bibitem[Haarnoja et~al\mbox{.}(2018a)]%
        {sac}
\bibfield{author}{\bibinfo{person}{Tuomas Haarnoja}, \bibinfo{person}{Aurick Zhou}, \bibinfo{person}{Pieter Abbeel}, {and} \bibinfo{person}{Sergey Levine}.} \bibinfo{year}{2018}\natexlab{a}.
\newblock \showarticletitle{Soft Actor-Critic: Off-Policy Maximum Entropy Deep Reinforcement Learning with a Stochastic Actor}. In \bibinfo{booktitle}{\emph{International Conference on Machine Learning}}. \bibinfo{pages}{1856--1865}.
\newblock


\bibitem[Haarnoja et~al\mbox{.}(2018b)]%
        {haarnoja2018soft}
\bibfield{author}{\bibinfo{person}{Tuomas Haarnoja}, \bibinfo{person}{Aurick Zhou}, \bibinfo{person}{Kristian Hartikainen}, \bibinfo{person}{George Tucker}, \bibinfo{person}{Sehoon Ha}, \bibinfo{person}{Jie Tan}, \bibinfo{person}{Vikash Kumar}, \bibinfo{person}{Henry Zhu}, \bibinfo{person}{Abhishek Gupta}, \bibinfo{person}{Pieter Abbeel}, {et~al\mbox{.}}} \bibinfo{year}{2018}\natexlab{b}.
\newblock \showarticletitle{Soft actor-critic algorithms and applications}.
\newblock \bibinfo{journal}{\emph{ArXiv preprint}}  \bibinfo{volume}{abs/1812.05905} (\bibinfo{year}{2018}).
\newblock


\bibitem[Heess et~al\mbox{.}(2015)]%
        {svg}
\bibfield{author}{\bibinfo{person}{Nicolas Heess}, \bibinfo{person}{Gregory Wayne}, \bibinfo{person}{David Silver}, \bibinfo{person}{Timothy Lillicrap}, \bibinfo{person}{Tom Erez}, {and} \bibinfo{person}{Yuval Tassa}.} \bibinfo{year}{2015}\natexlab{}.
\newblock \showarticletitle{Learning continuous control policies by stochastic value gradients}. In \bibinfo{booktitle}{\emph{Advances in Neural Information Processing Systems}}. \bibinfo{pages}{2944--2952}.
\newblock


\bibitem[Huber(1964)]%
        {huber}
\bibfield{author}{\bibinfo{person}{Peter~J. Huber}.} \bibinfo{year}{1964}\natexlab{}.
\newblock \showarticletitle{Robust Estimation of a Location Parameter}.
\newblock \bibinfo{journal}{\emph{Annals of Mathematical Statistics}} \bibinfo{volume}{35}, \bibinfo{number}{1} (\bibinfo{year}{1964}), \bibinfo{pages}{492--518}.
\newblock


\bibitem[Jimmy Lei~Ba(2016)]%
        {layernorm}
\bibfield{author}{\bibinfo{person}{Geoffrey E.~Hinton Jimmy Lei~Ba, Jamie Ryan~Kiros}.} \bibinfo{year}{2016}\natexlab{}.
\newblock \showarticletitle{Layer Normalization}.
\newblock \bibinfo{journal}{\emph{arXiv preprint}}  \bibinfo{volume}{abs/1607.06450} (\bibinfo{year}{2016}).
\newblock


\bibitem[Levine(2018)]%
        {pgm-rl}
\bibfield{author}{\bibinfo{person}{Sergey Levine}.} \bibinfo{year}{2018}\natexlab{}.
\newblock \showarticletitle{Reinforcement learning and control as probabilistic inference: Tutorial and review}.
\newblock \bibinfo{journal}{\emph{ArXiv preprint}}  \bibinfo{volume}{abs/1805.00909} (\bibinfo{year}{2018}).
\newblock


\bibitem[Liu et~al\mbox{.}(2025)]%
        {liu2025inferencetimes}
\bibfield{author}{\bibinfo{person}{Zijun Liu}, \bibinfo{person}{Peiyi Wang}, \bibinfo{person}{Runxin Xu}, \bibinfo{person}{Shirong Ma}, \bibinfo{person}{Chong Ruan}, \bibinfo{person}{Peng Li}, \bibinfo{person}{Yang Liu}, {and} \bibinfo{person}{Yu Wu}.} \bibinfo{year}{2025}\natexlab{}.
\newblock \bibinfo{title}{Inference-Time Scaling for Generalist Reward Modeling}.
\newblock
\newblock
\showeprint[arxiv]{2504.02495}~[cs.CL]
\urldef\tempurl%
\url{https://arxiv.org/abs/2504.02495}
\showURL{%
\tempurl}


\bibitem[Luo et~al\mbox{.}(2022)]%
        {spl-dqn}
\bibfield{author}{\bibinfo{person}{Yudong Luo}, \bibinfo{person}{Guiliang Liu}, \bibinfo{person}{Haonan Duan}, \bibinfo{person}{Oliver Schulte}, {and} \bibinfo{person}{Pascal Poupart}.} \bibinfo{year}{2022}\natexlab{}.
\newblock \showarticletitle{Distributional Reinforcement Learning with Monotonic Splines}. In \bibinfo{booktitle}{\emph{International Conference on Learning Representations}}.
\newblock


\bibitem[Ma et~al\mbox{.}(2022)]%
        {Ma2022MeanSemivariancePO}
\bibfield{author}{\bibinfo{person}{Xiaoteng Ma}, \bibinfo{person}{Shuai Ma}, \bibinfo{person}{Li Xia}, {and} \bibinfo{person}{Qianchuan Zhao}.} \bibinfo{year}{2022}\natexlab{}.
\newblock \showarticletitle{Mean-Semivariance Policy Optimization via Risk-Averse Reinforcement Learning}.
\newblock \bibinfo{journal}{\emph{J. Artif. Intell. Res.}}  \bibinfo{volume}{75} (\bibinfo{year}{2022}), \bibinfo{pages}{569--595}.
\newblock


\bibitem[Ma et~al\mbox{.}(2021)]%
        {Ma2021ConservativeOD}
\bibfield{author}{\bibinfo{person}{Yecheng~Jason Ma}, \bibinfo{person}{Dinesh Jayaraman}, {and} \bibinfo{person}{Osbert Bastani}.} \bibinfo{year}{2021}\natexlab{}.
\newblock \showarticletitle{Conservative offline distributional reinforcement learning}. In \bibinfo{booktitle}{\emph{Proceedings of the 35th International Conference on Neural Information Processing Systems}} \emph{(\bibinfo{series}{NIPS '21})}. \bibinfo{publisher}{Curran Associates Inc.}, \bibinfo{address}{Red Hook, NY, USA}, Article \bibinfo{articleno}{1471}, \bibinfo{numpages}{13}~pages.
\newblock
\showISBNx{9781713845393}


\bibitem[Markowitz(1959)]%
        {markowitz1959portfolio}
\bibfield{author}{\bibinfo{person}{Harry~M Markowitz}.} \bibinfo{year}{1959}\natexlab{}.
\newblock \bibinfo{booktitle}{\emph{Portfolio Selection: Efficient Diversification of Investments}}.
\newblock \bibinfo{publisher}{John Wiley \& Sons}, \bibinfo{address}{New York}.
\newblock


\bibitem[Mnih et~al\mbox{.}(2016)]%
        {a3c}
\bibfield{author}{\bibinfo{person}{Volodymyr Mnih}, \bibinfo{person}{Adria~Puigdomenech Badia}, \bibinfo{person}{Mehdi Mirza}, \bibinfo{person}{Alex Graves}, \bibinfo{person}{Timothy Lillicrap}, \bibinfo{person}{Tim Harley}, \bibinfo{person}{David Silver}, {and} \bibinfo{person}{Koray Kavukcuoglu}.} \bibinfo{year}{2016}\natexlab{}.
\newblock \showarticletitle{Asynchronous methods for deep reinforcement learning}. In \bibinfo{booktitle}{\emph{International conference on machine learning}}. \bibinfo{pages}{1928--1937}.
\newblock


\bibitem[Mnih et~al\mbox{.}(2013)]%
        {mnih2013playing}
\bibfield{author}{\bibinfo{person}{Volodymyr Mnih}, \bibinfo{person}{Koray Kavukcuoglu}, \bibinfo{person}{David Silver}, \bibinfo{person}{Alex Graves~Ioannis Antonoglou}, \bibinfo{person}{Daan Wierstra}, {and} \bibinfo{person}{Martin Riedmiller}.} \bibinfo{year}{2013}\natexlab{}.
\newblock \showarticletitle{Playing Atari with Deep Reinforcement Learning}.
\newblock \bibinfo{journal}{\emph{ArXiv preprint}}  \bibinfo{volume}{abs/1312.5602} (\bibinfo{year}{2013}).
\newblock


\bibitem[Mnih et~al\mbox{.}(2015)]%
        {dqn}
\bibfield{author}{\bibinfo{person}{Volodymyr Mnih}, \bibinfo{person}{Koray Kavukcuoglu}, \bibinfo{person}{David Silver}, \bibinfo{person}{Andrei~A. Rusu}, \bibinfo{person}{Joel Veness}, \bibinfo{person}{Marc~G. Bellemare}, \bibinfo{person}{Alex Graves}, \bibinfo{person}{Martin Riedmiller}, \bibinfo{person}{Andreas~K. Fidjeland}, \bibinfo{person}{Georg Ostrovski}, {et~al\mbox{.}}} \bibinfo{year}{2015}\natexlab{}.
\newblock \showarticletitle{Human-level control through deep reinforcement learning}.
\newblock \bibinfo{journal}{\emph{Nature}} \bibinfo{volume}{518}, \bibinfo{number}{7540} (\bibinfo{year}{2015}), \bibinfo{pages}{529}.
\newblock


\bibitem[Morimura et~al\mbox{.}(2010)]%
        {morimura2010nonparametric}
\bibfield{author}{\bibinfo{person}{Tetsuro Morimura}, \bibinfo{person}{Masashi Sugiyama}, \bibinfo{person}{Hisashi Kashima}, \bibinfo{person}{Hirotaka Hachiya}, {and} \bibinfo{person}{Toshiyuki Tanaka}.} \bibinfo{year}{2010}\natexlab{}.
\newblock \showarticletitle{Nonparametric return distribution approximation for reinforcement learning}. In \bibinfo{booktitle}{\emph{International Conference on Machine Learning}}. \bibinfo{pages}{799--806}.
\newblock


\bibitem[Nachum et~al\mbox{.}(2017)]%
        {nachum2017bridging}
\bibfield{author}{\bibinfo{person}{Ofir Nachum}, \bibinfo{person}{Mohammad Norouzi}, \bibinfo{person}{Kelvin Xu}, {and} \bibinfo{person}{Dale Schuurmans}.} \bibinfo{year}{2017}\natexlab{}.
\newblock \showarticletitle{Bridging the gap between value and policy based reinforcement learning}. In \bibinfo{booktitle}{\emph{Proceedings of the 31st International Conference on Neural Information Processing Systems}} (Long Beach, California, USA) \emph{(\bibinfo{series}{NIPS'17})}. \bibinfo{publisher}{Curran Associates Inc.}, \bibinfo{address}{Red Hook, NY, USA}, \bibinfo{pages}{2772–2782}.
\newblock
\showISBNx{9781510860964}


\bibitem[Ogryczak and Ruszczyński(1999)]%
        {ogryczak1999stochastic}
\bibfield{author}{\bibinfo{person}{Włodzimierz Ogryczak} {and} \bibinfo{person}{Andrzej Ruszczyński}.} \bibinfo{year}{1999}\natexlab{}.
\newblock \showarticletitle{From Stochastic Dominance to Mean-Risk Models: {{Semideviations}} as Risk Measures}.
\newblock \bibinfo{journal}{\emph{European Journal of Operational Research}} \bibinfo{volume}{116}, \bibinfo{number}{1} (\bibinfo{year}{1999}), \bibinfo{pages}{33--50}.
\newblock
\urldef\tempurl%
\url{https://doi.org/10.1016/S0377-2217(98)00167-2}
\showDOI{\tempurl}


\bibitem[Paszke et~al\mbox{.}(2019)]%
        {pytorch}
\bibfield{author}{\bibinfo{person}{Adam Paszke}, \bibinfo{person}{Sam Gross}, \bibinfo{person}{Francisco Massa}, \bibinfo{person}{Adam Lerer}, \bibinfo{person}{James Bradbury}, \bibinfo{person}{Gregory Chanan}, \bibinfo{person}{Trevor Killeen}, \bibinfo{person}{Zeming Lin}, \bibinfo{person}{Natalia Gimelshein}, \bibinfo{person}{Luca Antiga}, {et~al\mbox{.}}} \bibinfo{year}{2019}\natexlab{}.
\newblock \showarticletitle{{PyTorch}: An imperative style, high-performance deep learning library}. In \bibinfo{booktitle}{\emph{Advances in Neural Information Processing Systems}}. \bibinfo{pages}{8024--8035}.
\newblock


\bibitem[Prashanth et~al\mbox{.}(2022)]%
        {fu2018risk}
\bibfield{author}{\bibinfo{person}{LA Prashanth}, \bibinfo{person}{Michael~C Fu}, {et~al\mbox{.}}} \bibinfo{year}{2022}\natexlab{}.
\newblock \showarticletitle{Risk-Sensitive Reinforcement Learning via Policy Gradient Search}.
\newblock \bibinfo{journal}{\emph{Foundations and Trends{\textregistered} in Machine Learning}} \bibinfo{volume}{15}, \bibinfo{number}{5} (\bibinfo{year}{2022}), \bibinfo{pages}{537--693}.
\newblock


\bibitem[Prashanth and Ghavamzadeh(2016)]%
        {prashanth2016variance-constrained}
\bibfield{author}{\bibinfo{person}{L.~A. Prashanth} {and} \bibinfo{person}{Mohammad Ghavamzadeh}.} \bibinfo{year}{2016}\natexlab{}.
\newblock \showarticletitle{Variance-constrained actor-critic algorithms for discounted and average reward MDPs}.
\newblock \bibinfo{journal}{\emph{Machine Learning}} \bibinfo{volume}{105}, \bibinfo{number}{3} (\bibinfo{year}{2016}), \bibinfo{pages}{367--417}.
\newblock


\bibitem[Puterman(1994)]%
        {Puterman1994Markov}
\bibfield{author}{\bibinfo{person}{Martin~L. Puterman}.} \bibinfo{year}{1994}\natexlab{}.
\newblock \bibinfo{booktitle}{\emph{Markov Decision Processes: Discrete Stochastic Dynamic Programming}}.
\newblock \bibinfo{publisher}{John Wiley \& Sons, Inc.}
\newblock


\bibitem[Qu et~al\mbox{.}(2019)]%
        {qu2019nonlinear}
\bibfield{author}{\bibinfo{person}{Chao Qu}, \bibinfo{person}{Shie Mannor}, {and} \bibinfo{person}{Huan Xu}.} \bibinfo{year}{2019}\natexlab{}.
\newblock \showarticletitle{Nonlinear distributional gradient temporal-difference learning}. In \bibinfo{booktitle}{\emph{International Conference on Machine Learning}}. \bibinfo{pages}{5251--5260}.
\newblock


\bibitem[Rawlik et~al\mbox{.}(2013)]%
        {rawlik2013stochastic}
\bibfield{author}{\bibinfo{person}{Konrad Rawlik}, \bibinfo{person}{Marc Toussaint}, {and} \bibinfo{person}{Sethu Vijayakumar}.} \bibinfo{year}{2013}\natexlab{}.
\newblock \showarticletitle{On stochastic optimal control and reinforcement learning by approximate inference}. In \bibinfo{booktitle}{\emph{Twenty-Third International Joint Conference on Artificial Intelligence}}.
\newblock


\bibitem[Rockafellar and Uryasev(2000)]%
        {rockafellar2000optimization}
\bibfield{author}{\bibinfo{person}{R.~Tyrrell Rockafellar} {and} \bibinfo{person}{Stanislav Uryasev}.} \bibinfo{year}{2000}\natexlab{}.
\newblock \showarticletitle{Optimization of Conditional Value-at-Risk}.
\newblock \bibinfo{journal}{\emph{The Journal of Risk}} \bibinfo{volume}{2}, \bibinfo{number}{3} (\bibinfo{year}{2000}), \bibinfo{pages}{21--41}.
\newblock
\urldef\tempurl%
\url{https://doi.org/10.21314/JOR.2000.038}
\showDOI{\tempurl}


\bibitem[Rowland et~al\mbox{.}(2018a)]%
        {rowland2018analysis}
\bibfield{author}{\bibinfo{person}{Mark Rowland}, \bibinfo{person}{Marc Bellemare}, \bibinfo{person}{Will Dabney}, \bibinfo{person}{Remi Munos}, {and} \bibinfo{person}{Yee~Whye Teh}.} \bibinfo{year}{2018}\natexlab{a}.
\newblock \showarticletitle{An Analysis of Categorical Distributional Reinforcement Learning}. In \bibinfo{booktitle}{\emph{International Conference on Artificial Intelligence and Statistics}}. \bibinfo{pages}{29--37}.
\newblock


\bibitem[Rowland et~al\mbox{.}(2018b)]%
        {Rowland2018AnAnalysisCDRL}
\bibfield{author}{\bibinfo{person}{Mark Rowland}, \bibinfo{person}{Marc~G. Bellemare}, \bibinfo{person}{Will Dabney}, \bibinfo{person}{R{\'e}mi Munos}, {and} \bibinfo{person}{Yee~Whye Teh}.} \bibinfo{year}{2018}\natexlab{b}.
\newblock \showarticletitle{An Analysis of Categorical Distributional Reinforcement Learning}. In \bibinfo{booktitle}{\emph{International Conference on Artificial Intelligence and Statistics}}.
\newblock
\urldef\tempurl%
\url{https://api.semanticscholar.org/CorpusID:4861830}
\showURL{%
\tempurl}


\bibitem[Rowland et~al\mbox{.}(2019)]%
        {edrl}
\bibfield{author}{\bibinfo{person}{Mark Rowland}, \bibinfo{person}{Robert Dadashi}, \bibinfo{person}{Saurabh Kumar}, \bibinfo{person}{Remi Munos}, \bibinfo{person}{Marc~G Bellemare}, {and} \bibinfo{person}{Will Dabney}.} \bibinfo{year}{2019}\natexlab{}.
\newblock \showarticletitle{Statistics and Samples in Distributional Reinforcement Learning}. In \bibinfo{booktitle}{\emph{International Conference on Machine Learning}}. \bibinfo{pages}{5528--5536}.
\newblock


\bibitem[Schulman et~al\mbox{.}(2017a)]%
        {schulman2017equivalence}
\bibfield{author}{\bibinfo{person}{John Schulman}, \bibinfo{person}{Xi Chen}, {and} \bibinfo{person}{Pieter Abbeel}.} \bibinfo{year}{2017}\natexlab{a}.
\newblock \showarticletitle{Equivalence between policy gradients and soft q-learning}.
\newblock \bibinfo{journal}{\emph{ArXiv preprint}}  \bibinfo{volume}{abs/1704.06440} (\bibinfo{year}{2017}).
\newblock


\bibitem[Schulman et~al\mbox{.}(2015)]%
        {trpo}
\bibfield{author}{\bibinfo{person}{John Schulman}, \bibinfo{person}{Sergey Levine}, \bibinfo{person}{Pieter Abbeel}, \bibinfo{person}{Michael Jordan}, {and} \bibinfo{person}{Philipp Moritz}.} \bibinfo{year}{2015}\natexlab{}.
\newblock \showarticletitle{Trust region policy optimization}. In \bibinfo{booktitle}{\emph{International conference on machine learning}}. \bibinfo{pages}{1889--1897}.
\newblock


\bibitem[Schulman et~al\mbox{.}(2017b)]%
        {ppo}
\bibfield{author}{\bibinfo{person}{John Schulman}, \bibinfo{person}{Filip Wolski}, \bibinfo{person}{Prafulla Dhariwal}, \bibinfo{person}{Alec Radford}, {and} \bibinfo{person}{Oleg Klimov}.} \bibinfo{year}{2017}\natexlab{b}.
\newblock \showarticletitle{Proximal policy optimization algorithms}.
\newblock \bibinfo{journal}{\emph{ArXiv preprint}}  \bibinfo{volume}{abs/1707.06347} (\bibinfo{year}{2017}).
\newblock
\urldef\tempurl%
\url{https://arxiv.org/abs/1707.06347}
\showURL{%
\tempurl}


\bibitem[Shen et~al\mbox{.}(2014)]%
        {shen2014risk-sensitive}
\bibfield{author}{\bibinfo{person}{Yun Shen}, \bibinfo{person}{Michael~J. Tobia}, \bibinfo{person}{Tobias Sommer}, {and} \bibinfo{person}{Klaus Obermayer}.} \bibinfo{year}{2014}\natexlab{}.
\newblock \showarticletitle{Risk-sensitive reinforcement learning}.
\newblock \bibinfo{journal}{\emph{Neural Computation}} \bibinfo{volume}{26}, \bibinfo{number}{7} (\bibinfo{year}{2014}), \bibinfo{pages}{1298--1328}.
\newblock


\bibitem[Silver et~al\mbox{.}(2016)]%
        {silver2016mastering}
\bibfield{author}{\bibinfo{person}{David Silver}, \bibinfo{person}{Aja Huang}, \bibinfo{person}{Chris~J Maddison}, \bibinfo{person}{Arthur Guez}, \bibinfo{person}{Laurent Sifre}, \bibinfo{person}{George Van Den~Driessche}, \bibinfo{person}{Julian Schrittwieser}, \bibinfo{person}{Ioannis Antonoglou}, \bibinfo{person}{Veda Panneershelvam}, \bibinfo{person}{Marc Lanctot}, {et~al\mbox{.}}} \bibinfo{year}{2016}\natexlab{}.
\newblock \showarticletitle{Mastering the game of Go with deep neural networks and tree search}.
\newblock \bibinfo{journal}{\emph{nature}} \bibinfo{volume}{529}, \bibinfo{number}{7587} (\bibinfo{year}{2016}), \bibinfo{pages}{484}.
\newblock


\bibitem[Singh et~al\mbox{.}(2022)]%
        {sdpg}
\bibfield{author}{\bibinfo{person}{Rahul Singh}, \bibinfo{person}{Keuntaek Lee}, {and} \bibinfo{person}{Yongxin Chen}.} \bibinfo{year}{2022}\natexlab{}.
\newblock \showarticletitle{Sample-based distributional policy gradient}. In \bibinfo{booktitle}{\emph{Learning for Dynamics and Control Conference}}. PMLR, \bibinfo{pages}{676--688}.
\newblock


\bibitem[Singh et~al\mbox{.}(2020)]%
        {singh2020improving}
\bibfield{author}{\bibinfo{person}{Rahul Singh}, \bibinfo{person}{Qinsheng Zhang}, {and} \bibinfo{person}{Yongxin Chen}.} \bibinfo{year}{2020}\natexlab{}.
\newblock \showarticletitle{Improving robustness via risk averse distributional reinforcement learning}. In \bibinfo{booktitle}{\emph{Learning for Dynamics and Control}}. PMLR, \bibinfo{pages}{958--968}.
\newblock


\bibitem[Sobel(1982)]%
        {sobel1982variance}
\bibfield{author}{\bibinfo{person}{Matthew~J. Sobel}.} \bibinfo{year}{1982}\natexlab{}.
\newblock \showarticletitle{The variance of discounted Markov decision processes}.
\newblock \bibinfo{journal}{\emph{Journal of Applied Probability}} \bibinfo{volume}{19}, \bibinfo{number}{4} (\bibinfo{year}{1982}), \bibinfo{pages}{794--802}.
\newblock


\bibitem[Stooke and Abbeel(2019)]%
        {rlpyt}
\bibfield{author}{\bibinfo{person}{Adam Stooke} {and} \bibinfo{person}{Pieter Abbeel}.} \bibinfo{year}{2019}\natexlab{}.
\newblock \showarticletitle{rlpyt: A Research Code Base for Deep Reinforcement Learning in PyTorch}.
\newblock \bibinfo{journal}{\emph{ArXiv preprint}}  \bibinfo{volume}{abs/1909.01500} (\bibinfo{year}{2019}).
\newblock


\bibitem[Sun et~al\mbox{.}(2022)]%
        {Sun2022InterpretingDR}
\bibfield{author}{\bibinfo{person}{Ke Sun}, \bibinfo{person}{Yingnan Zhao}, \bibinfo{person}{Yi Liu}, \bibinfo{person}{Enze Shi}, \bibinfo{person}{Yafei Wang}, \bibinfo{person}{Aref Sadeghi}, \bibinfo{person}{Xiaodong Yan}, \bibinfo{person}{Bei Jiang}, {and} \bibinfo{person}{Linglong Kong}.} \bibinfo{year}{2022}\natexlab{}.
\newblock \showarticletitle{Interpreting Distributional Reinforcement Learning: Regularization and Optimization Perspectives}.
\newblock
\urldef\tempurl%
\url{https://api.semanticscholar.org/CorpusID:246473235}
\showURL{%
\tempurl}


\bibitem[Sutton and Barto(2018)]%
        {sutton2018reinforcement}
\bibfield{author}{\bibinfo{person}{Richard~S. Sutton} {and} \bibinfo{person}{Andrew~G. Barto}.} \bibinfo{year}{2018}\natexlab{}.
\newblock \bibinfo{booktitle}{\emph{Reinforcement learning: An introduction, Second Edition}}.
\newblock \bibinfo{publisher}{MIT press}.
\newblock


\bibitem[Tamar et~al\mbox{.}(2012)]%
        {castro2012policy}
\bibfield{author}{\bibinfo{person}{Aviv Tamar}, \bibinfo{person}{Dotan Di~Castro}, {and} \bibinfo{person}{Shie Mannor}.} \bibinfo{year}{2012}\natexlab{}.
\newblock \showarticletitle{Policy gradients with variance related risk criteria}. In \bibinfo{booktitle}{\emph{Proceedings of the 29th International Coference on International Conference on Machine Learning}}. \bibinfo{pages}{1651--1658}.
\newblock


\bibitem[Todorov(2008)]%
        {todorov2008general}
\bibfield{author}{\bibinfo{person}{Emanuel Todorov}.} \bibinfo{year}{2008}\natexlab{}.
\newblock \showarticletitle{General duality between optimal control and estimation}. In \bibinfo{booktitle}{\emph{2008 47th IEEE Conference on Decision and Control}}. IEEE, \bibinfo{pages}{4286--4292}.
\newblock


\bibitem[Todorov et~al\mbox{.}(2012)]%
        {mujoco}
\bibfield{author}{\bibinfo{person}{Emanuel Todorov}, \bibinfo{person}{Tom Erez}, {and} \bibinfo{person}{Yuval Tassa}.} \bibinfo{year}{2012}\natexlab{}.
\newblock \showarticletitle{MuJoCo: A physics engine for model-based control}. In \bibinfo{booktitle}{\emph{Intelligent Robots and Systems (IROS), 2012 IEEE/RSJ International Conference on}}.
\newblock


\bibitem[Tversky and Kahneman(1992)]%
        {cpt}
\bibfield{author}{\bibinfo{person}{Amos Tversky} {and} \bibinfo{person}{Daniel Kahneman}.} \bibinfo{year}{1992}\natexlab{}.
\newblock \showarticletitle{Advances in prospect theory: Cumulative representation of uncertainty}.
\newblock \bibinfo{journal}{\emph{Journal of Risk and uncertainty}} \bibinfo{volume}{5}, \bibinfo{number}{4} (\bibinfo{year}{1992}), \bibinfo{pages}{297--323}.
\newblock


\bibitem[Van~Hasselt et~al\mbox{.}(2016)]%
        {ddpg}
\bibfield{author}{\bibinfo{person}{Hado Van~Hasselt}, \bibinfo{person}{Arthur Guez}, {and} \bibinfo{person}{David Silver}.} \bibinfo{year}{2016}\natexlab{}.
\newblock \showarticletitle{Deep reinforcement learning with double q-learning}. In \bibinfo{booktitle}{\emph{Thirtieth AAAI conference on artificial intelligence}}.
\newblock


\bibitem[Wang(2000)]%
        {wang}
\bibfield{author}{\bibinfo{person}{Shaun~S. Wang}.} \bibinfo{year}{2000}\natexlab{}.
\newblock \showarticletitle{A class of distortion operators for pricing financial and insurance risks}.
\newblock \bibinfo{journal}{\emph{Journal of risk and insurance}} \bibinfo{volume}{67}, \bibinfo{number}{1} (\bibinfo{year}{2000}), \bibinfo{pages}{15--36}.
\newblock


\bibitem[Xia(2016)]%
        {xia2016optimization}
\bibfield{author}{\bibinfo{person}{Li Xia}.} \bibinfo{year}{2016}\natexlab{}.
\newblock \showarticletitle{Optimization of Markov decision processes under the variance criterion}.
\newblock \bibinfo{journal}{\emph{Automatica}}  \bibinfo{volume}{73} (\bibinfo{year}{2016}), \bibinfo{pages}{269--278}.
\newblock


\bibitem[Yang et~al\mbox{.}(2019)]%
        {fqf}
\bibfield{author}{\bibinfo{person}{Derek Yang}, \bibinfo{person}{Li Zhao}, \bibinfo{person}{Zichuan Lin}, \bibinfo{person}{Tao Qin}, \bibinfo{person}{Jiang Bian}, {and} \bibinfo{person}{Tie-Yan Liu}.} \bibinfo{year}{2019}\natexlab{}.
\newblock \showarticletitle{Fully Parameterized Quantile Function for Distributional Reinforcement Learning}. In \bibinfo{booktitle}{\emph{Advances in Neural Information Processing Systems}}. \bibinfo{pages}{6190--6199}.
\newblock


\bibitem[Zhang and Yao(2019)]%
        {zhang2019quota}
\bibfield{author}{\bibinfo{person}{Shangtong Zhang} {and} \bibinfo{person}{Hengshuai Yao}.} \bibinfo{year}{2019}\natexlab{}.
\newblock \showarticletitle{QUOTA: The quantile option architecture for reinforcement learning}. In \bibinfo{booktitle}{\emph{Proceedings of the AAAI Conference on Artificial Intelligence}}, Vol.~\bibinfo{volume}{33}. \bibinfo{pages}{5797--5804}.
\newblock


\bibitem[Zhou et~al\mbox{.}(2020)]%
        {nc-qr-dqn}
\bibfield{author}{\bibinfo{person}{Fan Zhou}, \bibinfo{person}{Jianing Wang}, {and} \bibinfo{person}{Xingdong Feng}.} \bibinfo{year}{2020}\natexlab{}.
\newblock \showarticletitle{Non-crossing quantile regression for deep reinforcement learning}. In \bibinfo{booktitle}{\emph{Proceedings of the 34th International Conference on Neural Information Processing Systems}} (Vancouver, BC, Canada) \emph{(\bibinfo{series}{NIPS '20})}. \bibinfo{publisher}{Curran Associates Inc.}, \bibinfo{address}{Red Hook, NY, USA}, Article \bibinfo{articleno}{1334}, \bibinfo{numpages}{11}~pages.
\newblock
\showISBNx{9781713829546}


\bibitem[Zhou et~al\mbox{.}(2021)]%
        {ndqfn}
\bibfield{author}{\bibinfo{person}{Fan Zhou}, \bibinfo{person}{Zhoufan Zhu}, \bibinfo{person}{Qi Kuang}, {and} \bibinfo{person}{Liwen Zhang}.} \bibinfo{year}{2021}\natexlab{}.
\newblock \showarticletitle{Non-decreasing Quantile Function Network with Efficient Exploration for Distributional Reinforcement Learning}. In \bibinfo{booktitle}{\emph{Proceedings of the Thirtieth International Joint Conference on Artificial Intelligence, {IJCAI-21}}}, \bibfield{editor}{\bibinfo{person}{Zhi-Hua Zhou}} (Ed.). \bibinfo{publisher}{International Joint Conferences on Artificial Intelligence Organization}, \bibinfo{pages}{3455--3461}.
\newblock
\urldef\tempurl%
\url{https://doi.org/10.24963/ijcai.2021/476}
\showDOI{\tempurl}
\newblock
\shownote{Main Track}.


\bibitem[Zhou et~al\mbox{.}(2018)]%
        {zhou2018infinite}
\bibfield{author}{\bibinfo{person}{Zhengyuan Zhou}, \bibinfo{person}{M Bloem}, {and} \bibinfo{person}{N Bambos}.} \bibinfo{year}{2018}\natexlab{}.
\newblock \showarticletitle{Infinite Time Horizon Maximum Causal Entropy Inverse Reinforcement Learning}.
\newblock \bibinfo{journal}{\emph{IEEE Trans. Automat. Control}} \bibinfo{volume}{63}, \bibinfo{number}{9} (\bibinfo{year}{2018}), \bibinfo{pages}{2787--2802}.
\newblock


\bibitem[Ziebart et~al\mbox{.}(2008)]%
        {meirl}
\bibfield{author}{\bibinfo{person}{Brian~D. Ziebart}, \bibinfo{person}{Andrew~L. Maas}, \bibinfo{person}{J.~Andrew Bagnell}, {and} \bibinfo{person}{Anind~K. Dey}.} \bibinfo{year}{2008}\natexlab{}.
\newblock \showarticletitle{Maximum Entropy Inverse Reinforcement Learning}. In \bibinfo{booktitle}{\emph{Proceedings of the Twenty-Third AAAI Conference on Artificial Intelligence, AAAI 2008, Chicago, Illinois, USA, July 13-17, 2008}}.
\newblock


\end{thebibliography}
	
	\appendix

	\section{Proofs} \label{app:proof}
		
	Before our proofs, we first review several useful properties of the metric $d_p$:
	\begin{align*}
		d_p(aU,aV)   & \leq |a|d_p(U,V) \tag{P1}      \\
		d_p(A+U,A+V) & \leq d_p(U,V) \tag{P2}         \\
		d_p(AU,AV)   & \leq \|A\|_pd_p(U,V), \tag{P3} 
	\end{align*}
	where $a$ is a scale and $A$ is a random variable independent of $U, V$. An additional lemma is required, which is called \emph{Partition lemma} by~\citet{c51}:
	\begin{equation*}
		d_p(U,V) \leq\sum_i d_p(A_i U,A_i V), \tag{P4}
	\end{equation*}
	where $A_i(\omega)\in \{0,1\}$ and for any $\omega$ there is exactly one $A_i$ with $A_i(\omega)=1$.
		
	We introduce the following inequality to deal with the relationship between policies and their action-value functions. Consider $\pi_\xi \in \mathbb{R}^{|\mathcal{A}|}$ parameterized by the softmax transform of $\xi \in \mathbb{R}^{|\mathcal{A}|}$ such that
	\begin{equation*}
		\pi_\xi(i) = \frac{\exp{\xi(i)}}{\sum_j \exp{\xi(j)}}, 1 \leq i \leq |\mathcal{A}|.
	\end{equation*}
	For any two vectors $\pi_{\xi_1}$ and $\pi_{\xi_2}$, we have,
	\begin{equation*}
		\| \log \pi_{\xi_1} - \log \pi_{\xi_2} \|_\infty \leq 2\| \xi_1 - \xi_2 \|_\infty . \tag{S1}
	\end{equation*}
	See Equation 68 in~\cite{cen2020fast} for the proof.
		
	In addition, we define a new transition operator $P_\pi$ to simplify the notation in the later proofs:
	\begin{equation*}
		P_\pi Z(s,a):\overset{D}{=} Z(s', a'), s' \sim P(\cdot \mid s,a), a' \sim \pi(\cdot \mid s).    
	\end{equation*}
	Further, we denote $P^*=P_{\pi^*}$.
		
	\setcounter{lemma}{0}
	\setcounter{theorem}{0}
	\setcounter{proposition}{0}
		
	\subsection{Lemma 1}
		
	\begin{lemma}
		$\mathcal{T}_{DS}^\pi: \mathcal{Z} \to \mathcal{Z}$ is a $\gamma$-contraction in $\bar{d}_p$.
	\end{lemma}
		
	\begin{proof} Let $Z_1, Z_2 \in \mathcal{Z}$ denote two action-value distributions. For any $(s,a) \in \mathcal{S}\times \mathcal{A}$, we have
		\begin{align*}
			  & d_p \big( \mathcal{T}_{DS}^{\pi} Z_{1}(s, a), \mathcal{T}_{DS}^{\pi} Z_{2}(s, a) \big)                                                                                                \\ 
			  & \qquad = d_p \big( R(s, a) + \gamma  P_\pi \left[Z_1(s, a) - \alpha \log \pi( a' \mid s')\right], R(s, a) + \gamma P_\pi \left[Z_2(s, a) - \alpha \log \pi( a' \mid s') \right] \big) \\
			  & \qquad \overset{(\romannumeral1)}{\leq} d_p\left(\gamma P_\pi  Z_{1}(s', a'), \gamma P_\pi Z_{2}(s', a')\right)                                                                       \\
			  & \qquad \overset{(\romannumeral2)}{\leq} \gamma d_p\left( P_\pi Z_{1}(s', a'), P_\pi Z_{2}(s', a')\right)                                                                              \\
			  & \qquad \leq \gamma \sup _{s^{\prime}, a^{\prime}} d_p \left(Z_{1}\left(s^{\prime}, a^{\prime}\right), Z_{2}\left(s^{\prime}, a^{\prime}\right)\right)                                 
		\end{align*}
		where $(\romannumeral1)$ follows by P2, $(\romannumeral2)$ follows by P1.  
		By definition of $\bar{d}_p$, we have
		\begin{align*}
			\bar{d}_p\big(\mathcal{T}_{DS}^{\pi} Z_1, \mathcal{T}_{DS}^{\pi} Z_2\big) & = \sup_{s,a} d_p \big( \mathcal{T}_{DS}^{\pi} Z_1(s,a), \mathcal{T}_{DS}^{\pi} Z_2(s,a) \big)                                                  \\
			                                                                          & \leq \gamma \sup _{s^{\prime}, a^{\prime}} d_p \left(Z_{1}\left(s^{\prime}, a^{\prime}\right), Z_{2}\left(s^{\prime}, a^{\prime}\right)\right) \\
			                                                                          & = \gamma \bar{d}_p\left( Z_1,  Z_2\right).  
		\end{align*}
	\end{proof}
		
	\subsection{Lemma 2}
		
	\begin{lemma}[Distributional Soft Policy Evaluation] Let $Z_{k+1} := \mathcal{T}_{DS}^\pi Z_k$ with $Z_0 \in \mathcal{Z}$. The sequence $Z_k$ will converge to $Z^\pi$ as $k \to \infty.$
	\end{lemma}
		
	\begin{proof}
		As we have proved that $\mathcal{T}_{DS}^\pi$ is a $\gamma$-contraction, policy evaluation can be obtained by repeatedly applying $\mathcal{T}_{DS}^\pi$.
	\end{proof}

	\subsection{Lemma 3}
		
	\begin{lemma}[Distributional Soft Policy Improvement] Let $Q^\pi(s,a) := \mathbb{E} [Z^\pi(s,a)]$,  $\forall (s,a) \in \mathcal{S} \times \mathcal{A}$. With $\pi_{\mathrm{old}} \in \Pi$ and $\pi_{\mathrm{new}}$ as the solution of problem defined in Equation~\ref{equ:sac-ori}, we have $Q^{\pi_{\mathrm{old}}}(s,a) \leq Q^{\pi_{\mathrm{new}}}(s,a)$.
	\end{lemma} 
		
	\begin{proof} 
		The proof of this lemma has a similar idea to that of the soft policy improvement in \cite{sac}. 
				
		For any policy $\pi \in \Pi$ and its soft action-value distribution $Z^\pi$, we define the soft action-value by taking the expectation,
		\begin{equation*}
			Q^\pi(s,a) = \mathbb{E} [Z^\pi(s,a)] = \mathbb{E} [R(s,a)] + \gamma \mathbb{E}_{s' \sim P(\cdot \mid s,a), a' \sim \pi(\cdot \mid s')} \left[ Z^\pi(s',a')-\alpha \log \pi( a' \mid s')\right].
		\end{equation*}
		Let denote a policy $\pi_{\text{old}} \in \Pi$ and its soft action-value $Q^{\pi_{\text{old}}}$. We can obtain a new policy $\pi_{\text{new}} \in \Pi$ by solving the minimization problem defined in Equation~\ref{equ:sac-ori},
		\begin{align*}
			\pi_{\text {new}} \left(\cdot \mid s \right) = & \mathop{\arg \min} \limits_{\pi^{\prime} \in \Pi} \mathrm{D}_{\mathrm{KL}} \left(\pi^{\prime}\left(\cdot \mid s\right) \| \exp \left( Q^{\pi_{\text {old }}} \left(s, \cdot\right)/\alpha-\log \Delta^{\pi_{\text {old }}}\left(s\right)\right)\right). 
		\end{align*}
		Since $\pi_{\text{new}}$ is the solution to the minimization problem above, it must be
		\begin{align*}
			\mathbb{E}_{a \sim \pi_{\text{new}}(\cdot \mid s)} 
			  & \left[ \log \pi_{\text{new}} \left(a \mid s\right) - Q^{\pi_{\text {old }}} \left(s, a \right) /\alpha + \log \Delta^{\pi_{\text {old }}} \left(s\right) \right] 
			\\
			  & \leq \mathbb{E}_{a \sim \pi_{\text{old}}(\cdot \mid s)}                                                                                                          
			\left[ \log \pi_{\text{old}} \left(a \mid s\right) - Q^{\pi_{\text {old }}}  \left(s, a \right) /\alpha + \log \Delta^{\pi_{\text {old }}} \left(s\right) \right],
		\end{align*}
		which can be reduced as follows for partition function $\log \Delta^{\pi_{\text {old }}}$ is only related to $s$,
		\begin{equation*}
			\mathbb{E}_{a \sim \pi_{\text{new}}(\cdot \mid s)} 
			\left[ \alpha \log \pi_{\text{new}} \left(a \mid s\right) - Q^{\pi_{\text {old }}} \left(s, a \right) \right] \leq \mathbb{E}_{a \sim \pi_{\text{old}}(\cdot \mid s)} 
			\left[ \alpha \log \pi_{\text{old}} \left(a \mid s\right) - Q^{\pi_{\text {old }}} \left(s, a \right)  \right].
		\end{equation*}
		Thus, we expand the soft Bellman equation and have 
		\begin{align*}
			Q^{\pi_{\text {old}}}  \left(s_t, a_t \right) & = \mathbb{E} [R (s_t, a_t )] + \gamma \mathbb{E}_{\substack{s_{t+1} \sim P(\cdot \mid s_t,a_t)    \\ a_{t+1} \sim \pi_{\text{old}}(\cdot \mid s_{t+1})}} \left[ Q^{\pi_{\text {old}}} (s_{t+1}, a_{t+1}) -\alpha \log \pi_{\text{old}} \left(a_{t+1} \mid s_{t+1} \right)  \right] \\
			                                              & \leq \mathbb{E} [R (s_t, a_t )] + \gamma \mathbb{E}_{\substack{s_{t+1} \sim P(\cdot \mid s_t,a_t) \\ a_{t+1} \sim \pi_{\text{new }}(\cdot \mid s_{t+1})}} \left[ Q^{\pi_{\text {old}}} (s_{t+1}, a_{t+1}) - \alpha \log \pi_{\text{new}} \left(a_{t+1} \mid s_{t+1} \right)  \right] \\
			                                              & \vdots                                                                                            \\
			                                              & \leq Q^{\pi_{\text {new}}}  \left(s_t, a_t \right).                                               
		\end{align*}
	\end{proof}
		
	\subsection{Theorem 1}
	\begin{theorem}[Convergence in the control setting] 
		Let $Z_{k+1} := \mathcal{T}_{DS}^* Z_k$ with $Z_0 \in \mathcal{Z}$. The sequence $Z_k$ will converge to $Z^*$ as $k \to \infty.$
	\end{theorem} 
		
	To prove this lemma, we first prove some auxiliary lemmas as follows.
		
	\begin{lemma}
		Let $Z:\mathcal{A}\to \mathcal{Z}$ denote a distribution function that maps actions to return distributions. For two stochastic policy $\pi_1, \pi_2$, we denote $Z_i :\overset{D}{=} Z(a), a \sim \pi_i$. We have $d_p(Z_1, Z_2) \leq D_{\rm TV}(\pi_1 \| \pi_2)B$, where $D_{\rm TV}$ is the total variation divergence and $B:=\sup_{Z\in \mathcal{Z}} \|Z\|_\infty < \infty$.
	\end{lemma} 
	\begin{proof}
		First, we denote a new random variable $W:=\mathbbm{1}_{a_1=a_2}, a_i \sim \pi_i(\cdot)$, while similarly, $\bar{W} := \mathbbm{1}_{a_1\neq a_2}$. We have
		\begin{align*}
			d_p(Z_1,Z_2) & = d_p((W+\bar{W})Z_1,(W+\bar{W})Z_2)                                                 \\
			             & \overset{(\romannumeral1)}{\leq} d_p(W Z_1, W Z_2) +d_p(\bar{W}Z_1, \bar{W}Z_2)      \\
			             & \overset{(\romannumeral2)}{\leq} d_p(\bar{W}Z_1, \bar{W}Z_2)                         \\
			             & \overset{(\romannumeral3)}{\leq} \mathrm{Pr}(a_1\neq a_2)d_p(\bar{W}Z_1, \bar{W}Z_2) \\
			             & \overset{(\romannumeral4)}{\leq} D_{\rm TV}(\pi_1 \| \pi_2) B,                       \\
		\end{align*}
		where $(\romannumeral1)$ follows by P4, $(\romannumeral2)$ follows by the fact $d_p(Z(a), Z(a))=0$, $(\romannumeral3)$ follows by P3, and $(\romannumeral4)$ follows by the definitions of $D_{\rm TV}(\pi_1 \| \pi_2)$ and $ B$.
	\end{proof} 
		
	\begin{lemma}
		Let $\pi_k, Q_k$ denote the policy and the action-value function in $k$-th iteration. Similarly, let denote $\pi^*, Q^*$ in the optimal case. We have
		\begin{equation*}
			D_{\rm KL}(\pi_k(\cdot \mid s) \| \pi^*(\cdot \mid s)) \leq 2\left\| Q_k(s,a) - Q^*(s,a) \right\|_\infty/ \alpha.
		\end{equation*}
	\end{lemma} 
		
	\noindent \textit{Proof.}
	By the definition of $D_{\rm KL}$,
	\begin{align*}
		D_{\rm KL}(\pi_k(\cdot \mid s) \| \pi^*(\cdot \mid s)) & = \sum_a \pi_k(a|s) \left[ \log \pi_k(a|s) - \log \pi^*(a|s)\right]              \\
		                                                       & \leq \| \pi_k(a|s)\|_1 \left\| \log \pi_k(a|s) - \log \pi^*(a|s) \right\|_\infty \\
		                                                       & \leq 2\left\| Q_k(s,a) - Q^*(s,a) \right\|_\infty/ \alpha,                       
	\end{align*}
	where the last inequality follows by S1.
		
	\begin{proof}[Proof of Theorem 1] 
		\begin{align*}
			d_{p} & \left(Z_{k+1}(s, a), Z^*(s, a)\right) = d_{p}\left(\mathcal{T}^*_{DS} Z_{k}(s, a), \mathcal{T}^*_{DS} Z^*(s, a)\right)                                                                          \\
			      & = d_{p}\left( R(s,a) + \gamma P_{\pi_k}[Z_{k}(s', a') - \alpha \log\pi_k( a' \mid s')], R(s,a) + \gamma P^*[Z^*(s', a') - \alpha \log\pi^*( a' \mid s')]\right)                                 \\
			      & \overset{(\romannumeral1)}{\leq} \gamma d_{p}\left(P_{\pi_k}[Z_{k}(s', a') - \alpha \log\pi_k( a' \mid s')], P^*[Z^*(s', a') - \alpha \log\pi^*( a' \mid s')]\right)                            \\
			      & \overset{(\romannumeral2)}{\leq} \gamma \underbrace{d_{p}\left(P_{\pi_k}[Z_{k}(s', a') - \alpha \log\pi_k( a' \mid s')], P_{\pi_k}[Z^*(s', a') - \alpha \log\pi_k( a' \mid s')]\right)}_{E_1} + \\
			      & \qquad \qquad \gamma \underbrace{d_{p}\left(P_{\pi_k}[Z^*(s', a') - \alpha \log\pi_k( a' \mid s')], P_{\pi_k}[Z^*(s', a') - \alpha \log\pi^*( a' \mid s')]\right)}_{E_2} +                      \\
			      & \qquad \qquad \gamma \underbrace{d_{p}\left(P_{\pi_k}[Z^*(s', a') - \alpha \log\pi^*( a' \mid s')], P^*[Z^*(s', a') - \alpha \log\pi^*( a' \mid s')]\right)}_{E_3},                             
		\end{align*}
		where $(\romannumeral1)$ follows P1 and P2, and $(\romannumeral2)$ follows the triangle inequality as $d_p$ is a metric.
				
		Similar with the result of Lemma~\ref{lem:operator}, we have
		\begin{align*}
			E_1 & \leq \bar{d_p}({Z_k, Z^*}). \\
			E_2 & \overset{(\romannumeral1)}{\leq} \alpha d_p(\log \pi_k( a' \mid s'), \log \pi^*( a' \mid s'))         \\
			    & \overset{(\romannumeral2)}{\leq} \alpha \|\log \pi_k( a' \mid s') - \log \pi^*( a' \mid s') \|_\infty \\
			    & \overset{(\romannumeral3)}{\leq} 2 \|Q_k(s', a') - Q^*(s', a') \|_\infty,                             
		\end{align*}
		where $(\romannumeral1)$ follows by P1 and P2, and $(\romannumeral3)$ follows by S1.
		\begin{align*}
			E_3 & \overset{(\romannumeral1)}{\leq} D_{\rm TV}(\pi_k(\cdot \mid s) \| \pi^*(\cdot \mid s)) B      \\
			    & \overset{(\romannumeral2)}{\leq} B \sqrt{\left\| Q_k(s,a) - Q^*(s,a) \right\|_\infty/ \alpha}, 
		\end{align*}
		where $(\romannumeral1)$  follow our lemma above, and $(\romannumeral2)$ follows the fact that $D_{\rm TV}(p \| q) \leq \sqrt{D_{\rm KL}(p \| q)/2} $.
				
		Combine all the inequalities above, we have 
		\begin{align*}
			d_{p}\left(Z_{k+1}(s, a), Z^*(s, a)\right) & \leq \gamma \left[\bar{d_p}(Z_k, Z^*) + 2C_k + B\sqrt{C_k/\alpha} \right], 
		\end{align*}
		where $C_k=\max_{s',a'} \|Q_k(s', a') - Q^*(s', a')\|_\infty$. Since soft Bellman operator $\mathcal{T}_{S}^*$ is a $\gamma$-contraction, we have $C_k \leq \gamma^k C_0$.
		Thus,  we infer that
		\begin{align*}
			\bar{d_{p}}\left(Z_{k+i+1}, Z^*\right) & \leq \gamma^i \bar{d_p}(Z_k, Z^*) + \sum_{j=0}^i \gamma^j \left(2 C_{k+j} + B \sqrt{C_{k+j}/\alpha} \right) \\
			                                       & \leq \gamma^i B + \frac{1}{1-\gamma} \left(2C_k+B \sqrt{C_{k}/\alpha} \right).                              
		\end{align*}
		For any $\epsilon > 0$, we can take $k$ and $i$ large enough to make $\bar{d_{p}}\left(Z_{k+i+1}, Z^* \right) < \epsilon$. We thus have completed this proof.
	\end{proof}
		
	\begin{proposition}
		$\mathcal{T}_{DS}^*$ is not a contraction.
	\end{proposition}
	\begin{proof}
		Next, we consider the similar example in~\cite{c51} and analyze the effect of adding entropy into the distributional optimality operator.
				
		There are two states $s_1, s_2$ and a unique transition from $s_1$ to $s_2$, where $s_2$ has two actions leads to different return distributions. The distributions are shown in Table~\ref{tab:counterexample}. In the table, we use a tuple $(Pr(z), z)$ to express the probability of an atom and its value, and denote $\diamondsuit = 1 + e^{\frac{\epsilon}{\alpha}}$ in shorthand.
				
		By the definition of $\bar{d_1}$, we have
				
		\begin{align*}
			\bar{d_1}(Z, Z^*)                             & = d_1(Z(s_2,a_2), Z^*(s_2,a_2)) = 2 \epsilon,                                                                         \\
			\bar{d_1}(\mathcal{\mathcal{T}_{DS}^*}Z, Z^*) & = d_1(Z(s_1), Z^*(s_1)) = \epsilon + (1-\epsilon)\frac{e^{\frac{\epsilon}{\alpha}}-1}{1+e^{\frac{\epsilon}{\alpha}}}. 
		\end{align*}

		The size relationship between $\bar{d_1}(Z, Z^*)$ and $\bar{d_1}(\mathcal{\mathcal{T}_{DS}^*}Z, Z^*)$ depends on both $\epsilon$ and $\alpha$. When $\epsilon=0.1$ and $\alpha=1$, $\bar{d_1}(Z, Z^*) < \bar{d_1}(\mathcal{\mathcal{T}_{DS}^*}Z, Z^*)$. However, $\bar{d_1}(Z, Z^*)>\bar{d_1}(\mathcal{\mathcal{T}_{DS}^*}Z, Z^*)$ if $\epsilon=0.1$ and $\alpha=1$, which reveals $\mathcal{T}_{DS}^*$ is a non-expansion. The result shows that although there is a positive impact on convergence of the distributional optimality operator, introducing entropy in policy objective can not make it a contraction.
				
		\begin{table}[ht]
			\caption{A Counterexample of Contraction}
			\centering
			\begin{tiny}
				\begin{tabular}{l|l|l|l}
					\toprule
					-                               & $s_1$                                                                                                                                                                                                                                                                                                    & $s_2, a_1$ & $s_2, a_2$                                                    \\
					\midrule
					$Z^*$                           & $\left( \frac{e^{\frac{\epsilon}{\alpha}}}{2\diamondsuit} ,-1+\alpha \log \left( \diamondsuit\right)\right), \left( \frac{1}{\diamondsuit} ,\alpha \log \left( \diamondsuit\right)\right), \left( \frac{e^{\frac{\epsilon}{\alpha}}}{2\diamondsuit} ,1+\alpha \log \left( \diamondsuit\right)\right)$    & (1,0)      & $(\frac{1}{2}, \epsilon - 1), (\frac{1}{2}, \epsilon + 1)) $  \\
					$Z$                             & $\left( \frac{e^{\frac{\epsilon}{\alpha}}}{2\diamondsuit} ,-1+\alpha \log \left( \diamondsuit\right)\right), \left( \frac{1}{\diamondsuit} ,\alpha \log \left( \diamondsuit\right)\right), \left( \frac{e^{\frac{\epsilon}{\alpha}}}{2\diamondsuit} ,1+\alpha \log \left( \diamondsuit\right)\right)$    & (1,0)      & $(\frac{1}{2}, -\epsilon - 1), (\frac{1}{2}, -\epsilon + 1) $ \\
					$\mathcal{\mathcal{T}_{DS}^*}Z$ & $\left( \frac{1}{2\diamondsuit} ,-\epsilon-1+\alpha \log \left( \diamondsuit\right)\right), \left( \frac{e^{\frac{\epsilon}{\alpha}}}{\diamondsuit} ,-\epsilon+\alpha \log \left( \diamondsuit\right)\right), \left( \frac{1}{2\diamondsuit} ,-\epsilon+1+\alpha \log \left( \diamondsuit\right)\right)$ & (1,0)      & $(\frac{1}{2}, \epsilon - 1), (\frac{1}{2}, \epsilon + 1)) $  \\
					\bottomrule
				\end{tabular}
			\end{tiny}
			\label{tab:counterexample}
		\end{table}
	\end{proof}
		
	\section{Implementation Details}
	\label{app:quantile-ablation}
	\subsection{Quantile fraction Generation}
\begin{figure}[H]
		\begin{center}
			\centerline{\includegraphics[width=\linewidth]{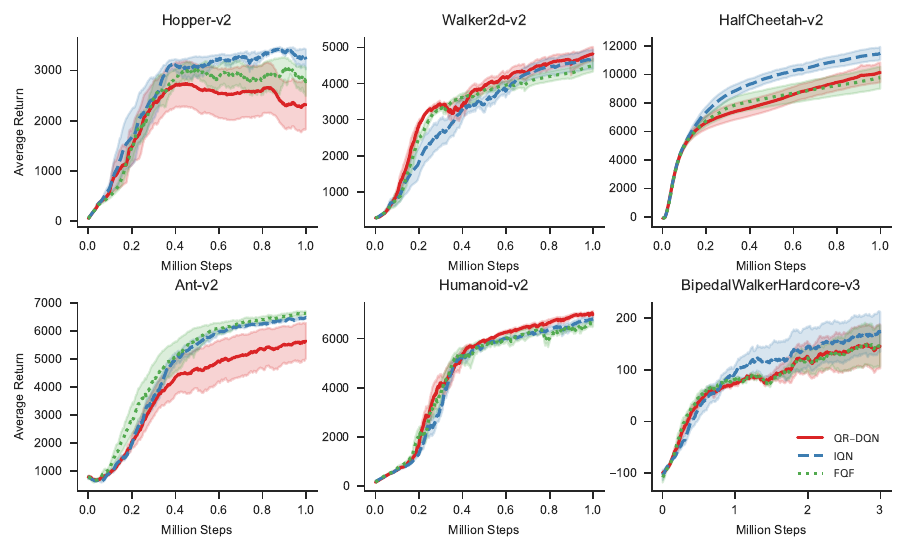}}
			\caption{Comparison of quantile fraction generation methods in MuJoCo and Box2d. Each learning curve is averaged over 5 different random seeds and shaded by the half of their variance. All curves are smoothed for better visibility.}
			\label{fig:ablation}
                \Description{Six line plots compare average return over training steps for three algorithms across different MuJoCo environments. The vertical axis shows average return and the horizontal axis shows training progress in millions of steps. Across most tasks, QR-DQN outperforms IQN and FQF, showing faster learning and higher final returns. The performance gap is most evident in Ant-v2 and BipedalWalkerHardcore-v3. Shaded regions indicate the variance over runs.}
		\end{center}
	\end{figure}

	We consider quantile regression for return distribution approximation in DSAC. Though we do not constraint which method to employ in quantile fraction generation, a modular experiment is provided. Out of six quantile function generation methods introduced in section~\ref{sec:intro} introduction, we adapt three earlier methods, QR-DQN, IQN and FQF, into DSAC and evaluate them on the MuJoCO and Box2d environments.  We show the results in Figure~\ref{fig:ablation}. Based on the experiment results, we choose random for quantile fraction generation, as it has better performance than fix and fewer parameters than net. The authors acknowledge our limitation that the later improved methods NC-QR-DQN, NDQFN and SPL-DQN were not included in this ablation studies.
		
	\textbf{Quantile Regression DQN (QR-DQN)} \citet{qr-dqn} first introduced quantile regression to improve the distribution approximation. By replacing the fix value atoms in C51 \cite{c51}, the value distribution is approximated by a group of trainable quantile values at fixed quantile fractions. With the basic idea of QR-DQN, quantile fractions are given by a group of fix values as $\tau_i=i / N, i=0, \ldots, N$.
		
	\textbf{Implicit Quantile Network (IQN)} \citet{iqn} extended the fixed quantile fractions to uniform samples and proposed implicit quantile value network (IQN). With infinite sampled quantile fractions, IQN is able to approximate the full quantile function. IQN uniformly samples the quantile fractions and takes the expectation in training. As we fix the number of quantile fractions and keep them in ascending order, we adapt the sampling as $\tau_0=0, \tau_i=\epsilon_i / \sum_{i=0}^{N-1} \epsilon_i$ where $\epsilon_i \sim U[0,1], i=1, \ldots, N$.
		
	\textbf{Fully parameterized Quantile Function (FQF)} Instead of sampling the quantile fractions uniformly, FQF parameterizes both the quantile values and quantile fractions. In addition to the quantile value network, FQF uses an additional fraction proposal network $q(s, a ; \omega)$, which has a softmax output layer and cumulatively sums the outputs as quantile fractions. The gradient of $\omega$ is given by the gradient of $W_1$ with respect to $\tau$,
		$\frac{\partial W_1}{\partial \tau_i}=2 Z_{\tau_i}(s, a ; \theta)-Z_{\hat{\tau}_i}(s, a ; \theta)-Z_{\hat{\tau}_{i-1}}(s, a ; \theta),$
		where $W_1$ denotes $1$-Wasserstein error between the approximated and true quantile functions. The quantile proposal network in FQF is a two-layer $128$-unit fully connected network with learning-rate set as 1e-5.

	\label{app:hyper-param}
	\subsection{Hyper-Parameter Setting} 
	Hyper-parameters used in experiments and environment specific parameters are listed in Table~\ref{tab:hyper1}.
		
	Instead of using fixed the temperature parameter $\alpha$ to balance reward and entropy, SAC has a variant~\shortcite{haarnoja2018soft} which introduces a mechanism of fine-tuning $\alpha$ to achieve target entropy adaptively. While our algorithm does not conflict with this parameter adaptive method, we use fixed $\alpha$ suggested in original SAC paper in our experiments to reduce irrelevant factors. For all experiments, we clip the rewards within $[-10, 10]$ to reduce the variation of value outputs, which is essential for BipedalWalkerHardcore-v3.

	\begin{table}[ht]
		\caption{Hyper-parameters - Environments}
		\label{tab:hyper2}
		\begin{center}
			\begin{tabular}{lcccr}
				\toprule
				Environment              & Temperature Parameter & Max episode length \\
				\midrule
				Hopper-v2                & 0.2                   & 1000               \\
				Walker2d-v2              & 0.2                   & 1000               \\
				HalfCheetah-v2           & 0.2                   & 1000               \\
				Ant-v2                   & 0.2                   & 1000               \\
				Humanoid-v2              & 0.05                  & 1000               \\
				BipedalWalkerHardcore-v3 & 0.01                  & 2000               \\
				Risky Mass Point         & 0.2                   & 100                \\
				Risky Ant                & 0.2                   & 200                \\
				\bottomrule
			\end{tabular}
		\end{center}
	\end{table}
		
	\begin{table}[ht]
		\caption{Hyper-parameters - Algorithms}
		\label{tab:hyper1}
		\begin{center}
			\begin{tabular}{lr}
				\toprule
				Hyper-parameter                                    & Value          \\
				\midrule
				\emph{Shared among all algorithms} \\
				\hspace{1em}Policy network learning rate           & 3e-4           \\
				\hspace{1em}(Quantile) Value network learning rate & 3e-4           \\
				\hspace{1em}Optimizer                              & Adam           \\
				\hspace{1em}Discount factor                        & 0.99           \\
				\hspace{1em}Target smoothing                       & 5e-3           \\
				\hspace{1em}Batch size                             & 256            \\
				\hspace{1em}Replay buffer size                     & 1e6            \\
				\hspace{1em}Minimum steps before training          & 1e4            \\
				\emph{Shared among TD3 \& TD4} \\
				\hspace{1em}Target policy noise                    & 0.2            \\
				\hspace{1em}Clip target policy noise               & 0.5            \\
				\hspace{1em}Policy update period                   & 2              \\
				\emph{Shared among DSAC \& TD4 \& SDPG} \\
				\hspace{1em}Number of quantile fractions           & 64             \\
				\hspace{1em}Quantile fraction embedding size       & 128            \\
				\hspace{1em}Huber regression threshold             & 1              \\
				\hspace{1em}Sample size                 & 100 \\
				\bottomrule
			\end{tabular}
		\end{center}
	\end{table}

	\subsection{Network Structure}
    	\begin{figure}[htbp]
		\centering
		\includegraphics[width=0.8\linewidth]{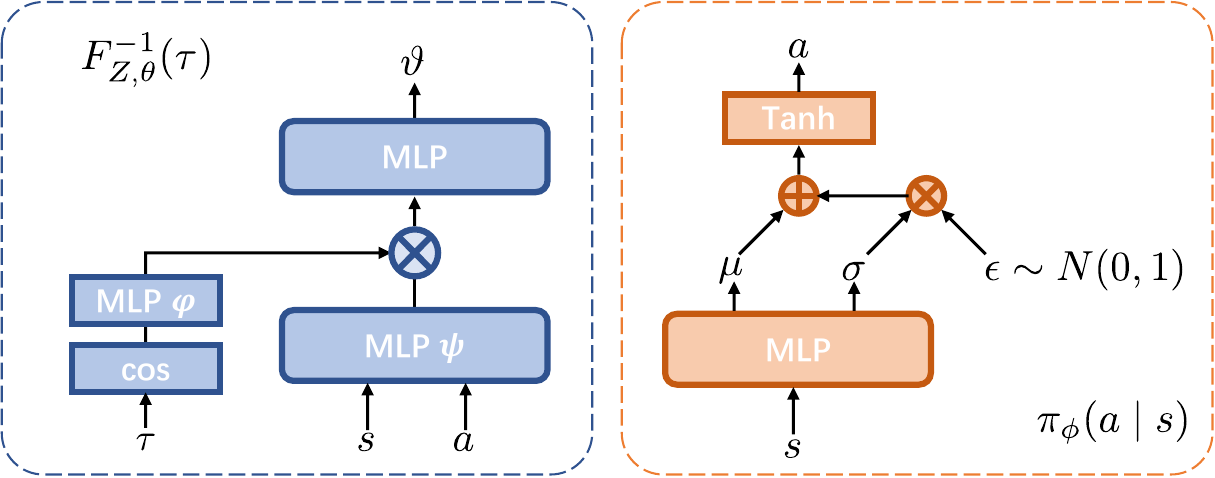}
		\caption{Network structures of DSAC.}
            \Description{Diagram shows the network architecture of the DSAC algorithm. The left plot depicts the quantile value network. A cosine embedding of the quantile input is processed by a small multi-layer perceptron and combined element-wise with the output of another MLP that encodes state and action. Their product is passed through a larger MLP to produce the quantile value. The right plot shows the stochastic policy network. The state is encoded by an MLP to output mean and standard deviation, which are used to sample actions with reparameterization and passed through a Tanh function.}
	\end{figure}
	
	For SAC and TD3, we utilize two-layer fully connected networks with 256 hidden units for both value and policy network and $\mathrm{ReLU}$ as activation function. DSAC has the same structure of policy network $\pi_\phi(a \mid s; \phi)$.
		
	In order to embed quantile fraction $\tau$ into the quantile value network, we borrow the idea of implicit representations from IQN~\cite{iqn}. With infinite sampled quantile fractions, IQN is able to approximate the full quantile function. IQN uniformly samples the quantile fractions and takes the expectation in training. As we fix the number of quantile fractions and keep them in ascending order, we adapt the sampling as $\tau_0 = 0, \tau_i = \epsilon_i/\sum_{i=0}^{N-1} \epsilon_i$ where $\epsilon_i \sim U[0,1] , i = 1, \dots, N$. With the element-wise (Hadamard) product (denoted as $\odot$) of state feature $\psi(s,a)$ and embedding $\varphi(\tau)$, the quantile values are given by $Z_\tau(s,a) = Z(\psi(s,a) \odot (1 + \varphi(\tau));\theta) $. After studying  different ways for embedding $\tau$, IQN suggests an embedding formula with $\cos(\cdot)$, 
	\begin{equation}
		\label{equ:tau-embed}
		\varphi_{j}(\tau):=\text{ReLU}\left(\sum_{i=0}^{n-1} \cos (i \pi \tau) w_{i 
			j}+b_{j}\right),
	\end{equation}
	where $n$ is the embedding dimension and $w_{i j},b_j$ are network parameters.
		
	We use one-layer 256-unit fully connected network for $\psi(s,a)$ and one-layer 64-unit fully connected network for $\varphi(\tau)$ in Equation~\ref{equ:tau-embed}, followed with one-layer 256-unit fully connected network for their multiplied values $\psi(s,a) \odot (1 + \varphi(\tau))$. Moreover, since magnitudes of quantile values vary hugely, we implement layer normalization~\cite{layernorm} to the hidden activation of networks.

	\subsection{Additional Results}
	\begin{figure}[H]
		\centering
		        
		\begin{minipage}[b]{0.32\linewidth}
			\centering
			\includegraphics[width=\linewidth]{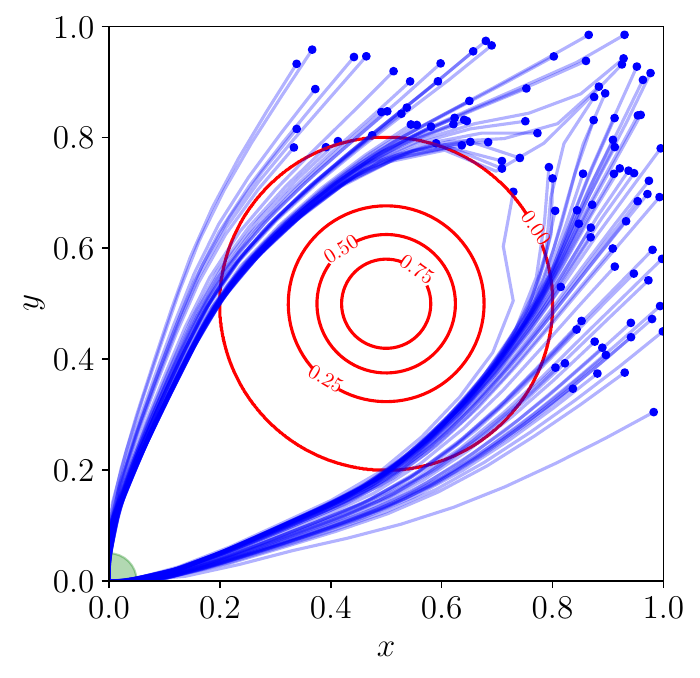}
			{\small (a) Wang(0.75)}
		\end{minipage}
		\hspace{0.01\linewidth}
		\begin{minipage}[b]{0.32\linewidth}
			\centering
			\includegraphics[width=\linewidth]{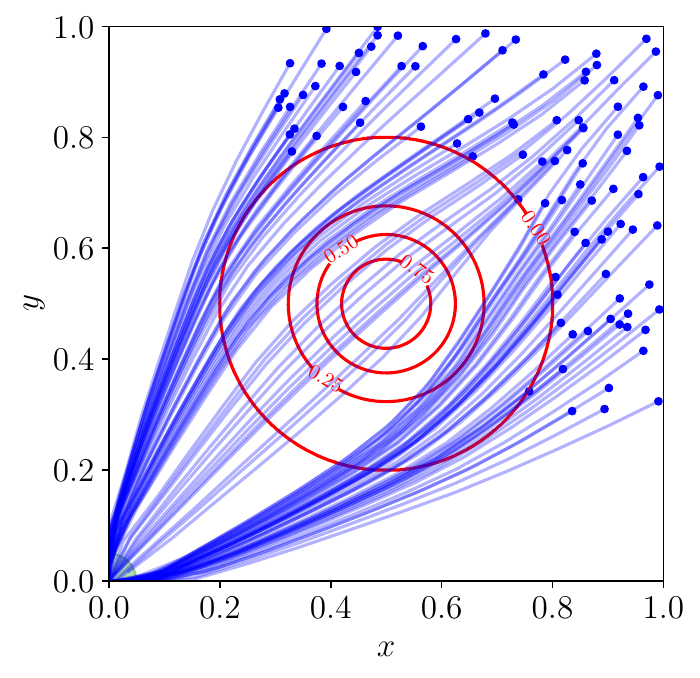}
			{\small (b) CPW(0.71)}
		\end{minipage}
		\hspace{0.01\linewidth}
		\begin{minipage}[b]{0.32\linewidth}
			\centering
			\includegraphics[width=\linewidth]{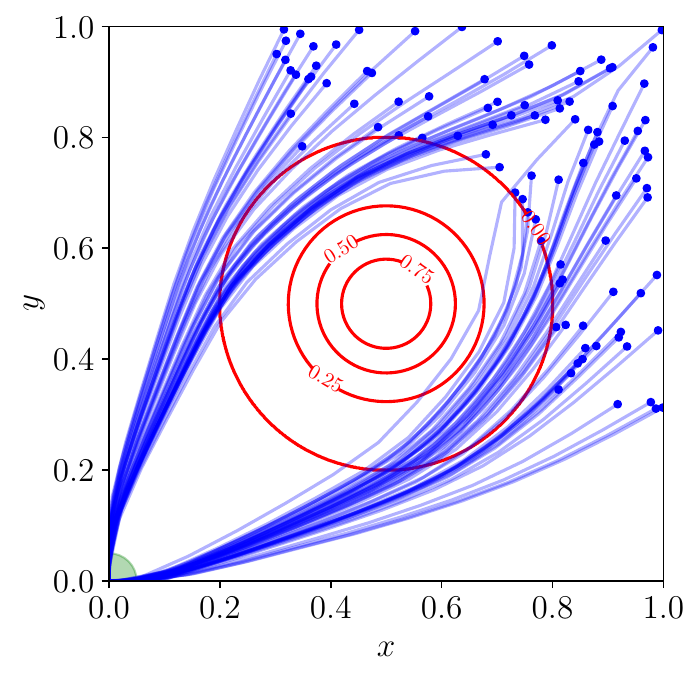}
			{\small (c) MSD(1)}
		\end{minipage}
		        
		\caption{Visualization of sampled trajectories in Risk Mass Point. For each risk measure, we choose its best seed in terms of average return.}
		\label{fig:risk_point_add}
            \Description{Three contour plots show two-dimensional trajectory distributions with red distortion function level sets and blue sample paths. The x-axis and y-axis range from zero to one, representing normalized space. Each plot shows multiple blue trajectories originating from the lower left corner, with end points marked as blue dots. The red concentric contours represent distortion function values from 0.25 to 1.00. In the first and third plots, most trajectories curve around and avoid the outer contours. In the second plot, more trajectories pass through the outer risk regions, meaning less effective avoidance under distortion.}
	\end{figure}
	\begin{figure}[H]
		\centering
		        
		\begin{minipage}[b]{0.32\linewidth}
			\centering
			\includegraphics[width=\linewidth]{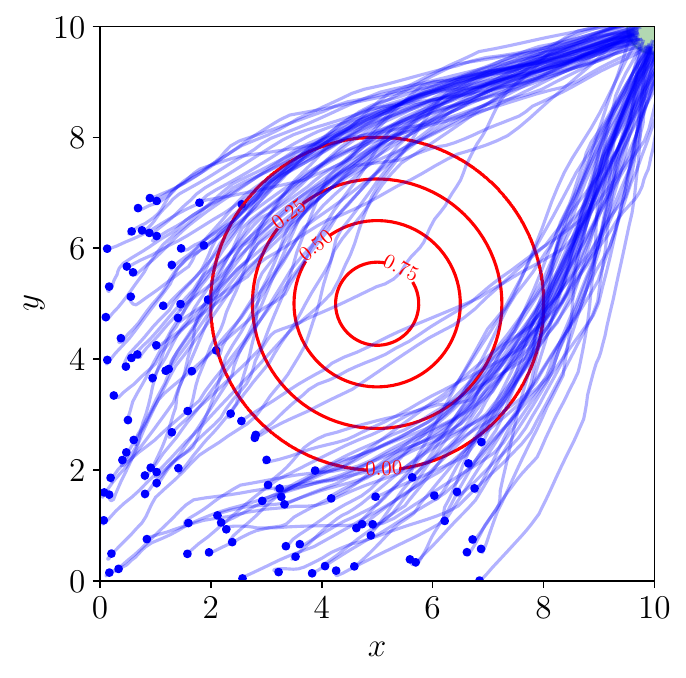}
			{\small (a) Wang(0.75)}
		\end{minipage}
		\hspace{0.01\linewidth}
		\begin{minipage}[b]{0.32\linewidth}
			\centering
			\includegraphics[width=\linewidth]{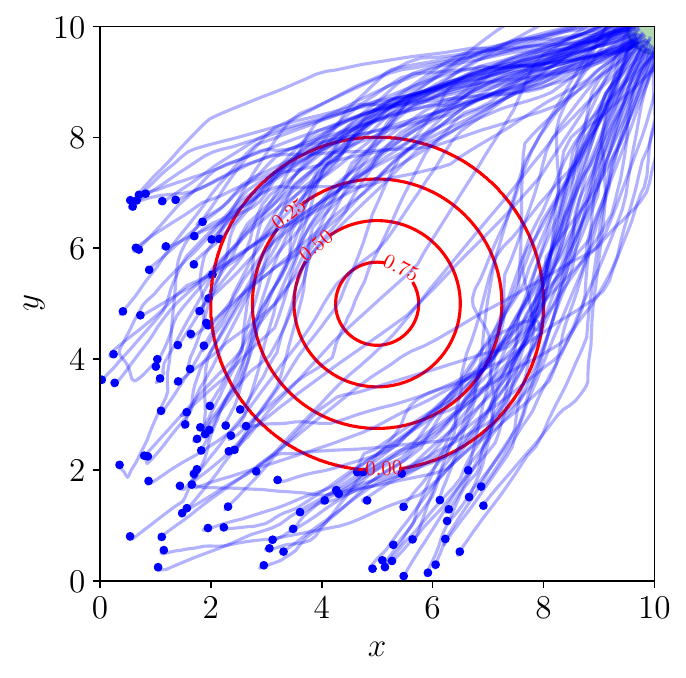}
			{\small (b) CPW(0.71)}
		\end{minipage}
		\hspace{0.01\linewidth}
		\begin{minipage}[b]{0.32\linewidth}
			\centering
			\includegraphics[width=\linewidth]{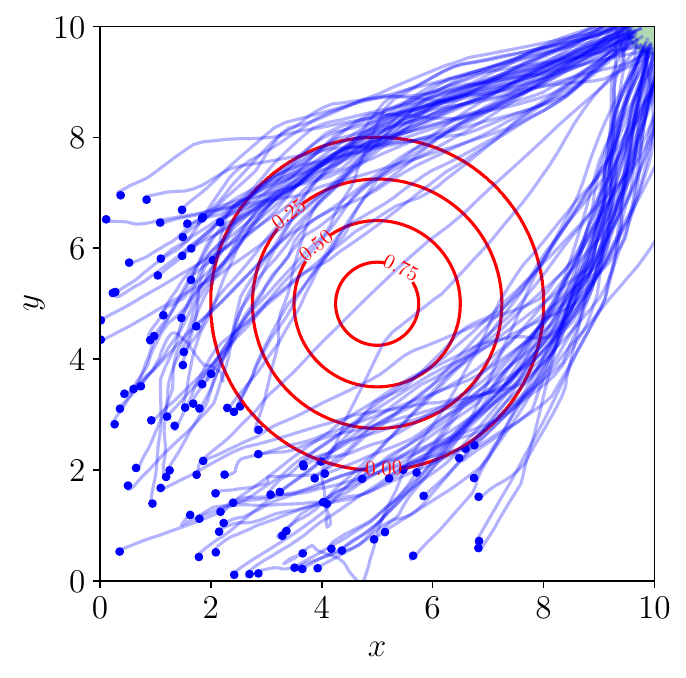}
			{\small (c) MSD(1)}
		\end{minipage}
		        
		\caption{Visualization of sampled trajectories in Risk Ant. For each risk measure, we choose its best seed in terms of average return.}
		\label{fig:risk_ant_add}
            \Description{Three contour plots show two-dimensional trajectory distributions in a 10 by 10 space with red distortion function level sets and blue sample paths. Blue trajectories start from the lower left and terminate near the upper right corner, with end points marked as blue dots. Red concentric contours represent distortion values from 0.25 to 1.00. In the first plot, several trajectories pass through the outer risk regions. The second plot shows moderate avoidance of the red contours. The third plot shows most trajectories bending away from the outer contours, meaning improved risk-aware behavior.}
	\end{figure}
	
	%%%%%%%%%%%%%%%%%%%%%%%%%%%%%% template %%%%%%%%%%%%%%%%%%%%%%%%%%%%%%%%%%%%%

\end{document}